\def\ARXIVVERSION{1}
\documentclass[11pt]{article}

\usepackage[preprint]{acl}

\usepackage{times}
\usepackage{latexsym}
\usepackage[T1]{fontenc}
\usepackage[utf8]{inputenc}
\usepackage{microtype}
\usepackage{inconsolata}
\usepackage{graphicx}
\usepackage{amsmath}
\usepackage{amssymb}
\usepackage{booktabs}
\usepackage{multirow}
\usepackage{tabularx}
\usepackage{algorithm}
\usepackage{algpseudocode}
\usepackage[most]{tcolorbox}

\newcolumntype{Y}{>{\centering\arraybackslash}X}

\tcbset{
  casebox/.style={
    enhanced,
    breakable,
    colback=gray!4,
    colframe=gray!55,
    boxrule=0.4pt,
    arc=2pt,
    left=6pt, right=6pt, top=4pt, bottom=4pt,
    fonttitle=\bfseries\small,
    coltitle=black,
    colbacktitle=gray!18,
    attach boxed title to top left={xshift=6pt, yshift=-2pt},
    boxed title style={colframe=gray!55, boxrule=0.4pt, arc=2pt},
  },
}

\title{LEAP: Likelihood Elicitation and Aggregation for LLM-based Probabilistic Forecasting}

\author{
  \textbf{Yufei Chen}$^{1}$ \quad
  \textbf{Yiran Zhao}$^{2}$ \quad
  \textbf{Xiaogang Xu}$^{3,4,\dagger}$
  \\
  \textbf{Qipeng Xie}$^{5}$ \quad
  \textbf{Jiafei Wu}$^{3,4}$ \quad
  \textbf{Zhe Liu}$^{3,4,\dagger}$
  \\[0.5em]
  {\normalsize\normalfont $^1$Shandong University \quad
  $^2$Nanjing University of Aeronautics and Astronautics}
  \\
  {\normalsize\normalfont $^3$School of Software Technology, Zhejiang University}
  \\
  {\normalsize\normalfont $^4$Ningbo Global Innovation Center, Zhejiang University}
  \\
  {\normalsize\normalfont $^5$The Hong Kong University of Science and Technology (Guangzhou)}
}

\begin{document}
\maketitle
\begingroup
\renewcommand{\thefootnote}{\fnsymbol{footnote}}
\footnotetext[2]{Corresponding authors: \texttt{xiaogangxu00@gmail.com} and \texttt{zhe.liu@zju.edu.cn}.}
\endgroup

\begin{abstract}
\sloppy
LLM-based forecasting systems have improved on real-world tasks such as financial markets and sports outcomes, largely through stronger search and tool use. Many systems still ask an LLM to read all collected evidence together and produce the final forecast. We call this design \emph{Monolithic Prediction}. It can obscure how individual evidence items affect the result and collapse uncertainty across competing outcomes. We propose \textbf{LEAP} (\textbf{L}ikelihood \textbf{E}licitation and \textbf{A}ggregation for \textbf{P}robabilistic forecasting), which reorganizes how collected evidence is used in the prediction stage. LEAP examines each evidence item separately and elicits likelihood parameters that describe its implications for the target. An explicit prior and a deterministic probabilistic model then combine these likelihoods into a posterior distribution. This procedure supports continuous, single-choice, and multi-choice forecasts while preserving reproducible evidence contributions. We build a benchmark covering forecasting, information-seeking, and browsing tasks, and evaluate LEAP on our own agent loop and several agent CLI frameworks. Given the same evidence, LEAP improves most prediction and calibration metrics across models and remains stronger under controlled comparisons of prior access, inference budget, and aggregation.\ifdefined\ARXIVVERSION\footnote{\url{https://github.com/layingfish/LEAP}}\fi

\end{abstract}

\begin{figure}[t]
\centering
\includegraphics[width=\columnwidth]{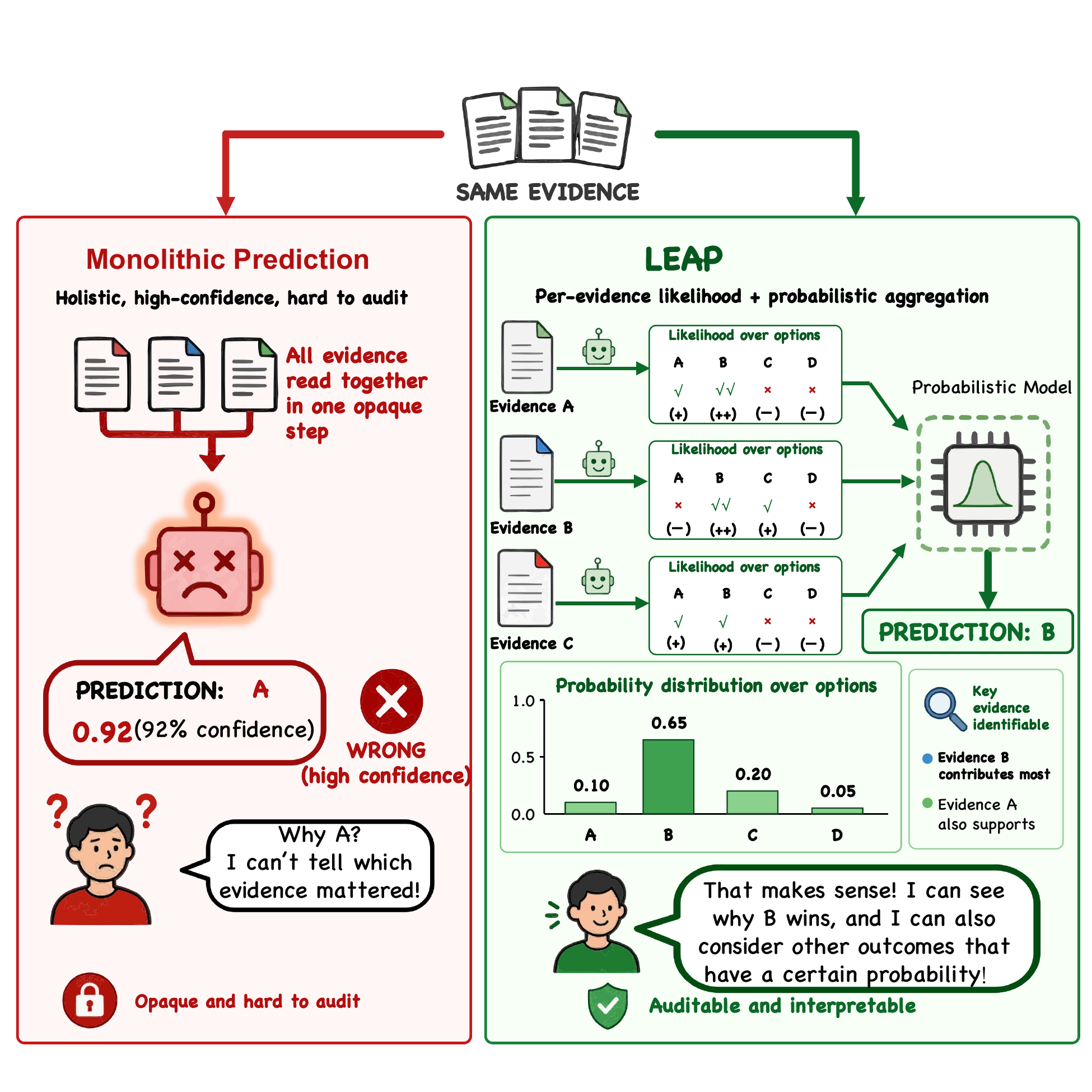}
\caption{The Monolithic baseline can be confidently wrong without revealing which evidence drove the answer. LEAP exposes evidence-level support and combines it through a probabilistic update, making the prediction auditable.}
\label{fig:motivation}
\end{figure}

\section{Introduction}

As large language models (LLMs) and agent harnesses mature, LLM-based agents are increasingly used for real-world forecasting, from economic and financial indicators to sports outcomes and geopolitical events \citep{zou2022forecasting,karger2025forecastbench,yang2025prophetarena,halawi2024approaching,schoenegger2024wisdom,alur2025aia}. A representative system first runs an agent loop that searches the web and other sources for relevant material, then passes the collected evidence to one or more LLM calls that read it as a whole and produce the final prediction. Recent work has improved the first stage through agentic search loops, retrieval, tool use, web interaction, and source integration \citep{yan2024autocastpp,nakano2021webgpt,yao2023react,schick2023toolformer,trivedi2023interleaving,qin2024toolllm,liu2024agentbench,zhou2024webarena,deng2023mind2web,xie2024osworld}. A common design for the second stage remains holistic: one or more LLM forecasting calls read everything gathered and write out the answer.

This final prediction stage has two undesirable properties. First, it is opaque: the user can read the LLM's rationale, but cannot isolate the influence of any specific evidence item on the result; as a result, the forecast may collapse evidence-supported uncertainty over competing outcomes into a single answer \citep{deyoung2020eraser,lyu2023faithful}. Second, because a point forecast requires one answer, the LLM must merge evidence from different sources into a single supporting context. In this long-context setting, the final prediction is more likely to be biased by LLM forgetting and hallucination \citep{liu2024lost,ji2023survey,niu2024ragtruth,belem2025single}. Figure~\ref{fig:motivation} illustrates this failure mode.

Probabilistic models provide a natural way to keep evidence interpretation separate from aggregation. BIRD uses LLM-generated factors and coarse probability judgments to parameterize a Bayesian model for probabilistic inference \citep{feng2025bird}. Nafar et al.\ elicit conditional probabilities from LLMs to parameterize predefined Bayesian networks \citep{nafar2025extracting}. Concurrent work on the Bayesian Linguistic Forecaster maintains a structured belief state during iterative search and combines multiple forecasts through shrinkage and calibration \citep{murphy2026agentic}. These studies demonstrate practical ways to place LLM judgments inside structured probabilistic models. We apply the same general idea after evidence collection, using an explicit Bayesian model to combine the gathered evidence into a forecast.

We therefore propose \textbf{LEAP} (\textbf{L}ikelihood \textbf{E}licitation and \textbf{A}ggregation for \textbf{P}robabilistic forecasting), which reorganizes how gathered information is consumed in the final prediction step. Rather than asking the LLM to forecast from the full evidence bundle, LEAP uses it as a local interpreter: the LLM examines each evidence item in isolation and reports likelihood parameters describing what the evidence implies about the target. Alongside these likelihoods, a prior over the answer is also constructed, supplying a base rate before any evidence is examined. A deterministic probabilistic model then combines this prior with the individual evidence likelihoods and returns a posterior distribution over the answer.

This design gives LEAP two useful properties. First, the output is a posterior distribution rather than a point estimate, improving prediction accuracy and giving human decision makers richer information for comparing competing outcomes. Second, every step is inspectable: the user can see which evidence items entered the posterior, what each contributed, and how the final distribution was assembled. A forecast can therefore be audited at the level of individual evidence items, for example by removing one item and recomputing the prediction.

We evaluate LEAP on a benchmark drawn from FutureX, GAIA, and BrowseComp \citep{zeng2025futurex,mialon2024gaia,wei2025browsecomp}, covering forecasting, information-seeking, and browsing tasks. To make the comparison fair, the evaluation uses strict temporal isolation, model selection constrained by knowledge cutoffs, and a shared evidence set for both methods (details in Section~\ref{sec:setup}). Under this setup, LEAP improves over \emph{Monolithic} on most settings and metrics, both on our custom agent loop across multiple base models and as a probability skill atop several agent CLI frameworks used in practice.

Our contributions are: \emph{(1)} we study how a fixed evidence set should be converted into a probabilistic forecast and isolate this prediction step from evidence collection; \emph{(2)} we propose LEAP, which uses the LLM as a local interpreter of each evidence item and aggregates its outputs through a deterministic probabilistic model into a posterior over the answer; \emph{(3)} we construct a benchmark covering forecasting, information-seeking, and browsing tasks on which LEAP improves prediction accuracy and probabilistic quality across multiple base models and several agent CLI frameworks; and \emph{(4)} we provide diagnostic analyses showing where the gains come from and how forecasts can be audited at the evidence-item level.

\section{Related Work}

\paragraph{LLM-based forecasting systems.}
LLM-based forecasting has moved from benchmark construction toward systems that combine search, reasoning, aggregation, and calibration. AutoCast introduced temporally grounded forecasting questions with timestamped evidence \citep{zou2022forecasting}; AutoCast++ improved world event prediction through zero-shot context retrieval and summarization \citep{yan2024autocastpp}; ForecastBench extends this line with a dynamic benchmark and human comparisons \citep{karger2025forecastbench}; and Prophet Arena studies the predictive intelligence of LLMs in forecasting settings \citep{yang2025prophetarena}. Concurrent work shows that ensembles of LLM forecasters can approach human crowd forecasts on binary questions \citep{schoenegger2024wisdom}. Recent systems give the LLM search access and combine decomposition, retrieval, ensembling, and calibration after prediction, but the final forecast is still produced by LLM calls that read the gathered material as a whole \citep{halawi2024approaching,alur2025aia}. We focus on what happens after evidence collection: how should fixed evidence be converted into the final prediction? Our method assesses what each evidence item implies about the candidate outcomes and keeps these local assessments separate until aggregation.

\paragraph{LLM-assisted Bayesian inference.}
BIRD uses LLM-generated factors and coarse probability judgments to parameterize a Bayesian model for probabilistic inference \citep{feng2025bird}. Nafar et al.\ elicit conditional probabilities from LLMs to parameterize predefined Bayesian networks, especially when observed data are limited \citep{nafar2025extracting}. The Bayesian Linguistic Forecaster maintains a linguistic belief state during iterative search and combines multiple forecasts through shrinkage and calibration \citep{murphy2026agentic}. These systems show how LLM judgments can support structured probabilistic inference.

\paragraph{Evidence collection and agent evaluation.}
Agent methods are directly relevant to the first stage of our setting: collecting and organizing evidence before the prediction is made. Retrieval-augmented generation and passage readers ground generation and open-domain question answering in external documents \citep{lewis2020retrieval,izacard2021leveraging}, retrieval can reduce hallucination in dialogue \citep{shuster2021retrieval}, WebGPT connects browser use with long-form question answering \citep{nakano2021webgpt}, and ReAct interleaves reasoning and acting \citep{yao2023react}. Related methods decompose questions, sample multiple reasoning paths, learn tool use, call large API collections, or reflect on previous attempts \citep{press2023measuring,wei2022chain,wang2023selfconsistency,schick2023toolformer,patil2024gorilla,qin2024toolllm,shinn2023reflexion}. Agent benchmarks then score the combined effect of search, reasoning, tool use, and answer generation, from multi-hop QA and interleaved retrieval to web, desktop, browsing, and forecasting tasks \citep{yang2018hotpotqa,trivedi2023interleaving,liu2024agentbench,zhou2024webarena,deng2023mind2web,xie2024osworld,mialon2024gaia,wei2025browsecomp,zou2022forecasting,karger2025forecastbench,zeng2025futurex}. These evaluations score whole systems; our protocol holds the gathered evidence fixed, isolating how that evidence is turned into a forecast.

\paragraph{Auditability and faithful explanations.}
Our focus on evidence-level auditability is related to work on rationales and faithful explanations in NLP. ERASER evaluates rationalized models using human-marked supporting evidence and faithfulness metrics \citep{deyoung2020eraser}. Faithful chain-of-thought work similarly argues that a reasoning trace is most useful when it is coupled to the computation that produces the answer, for example by delegating final execution to a deterministic solver \citep{lyu2023faithful}. LEAP follows this principle in forecasting: the LLM supplies local evidence interpretations, while the posterior and evidence contributions are computed by an explicit probabilistic update rather than by a simple explanation written after the fact.

\paragraph{Scope of LEAP.}
LEAP operates after an upstream system has collected evidence and focuses on the final prediction step. BIRD builds a Bayesian model from LLM-generated factors for a target query, while Nafar et al.\ parameterize predefined domain-level networks from LLM judgments \citep{feng2025bird,nafar2025extracting}. LEAP starts from retrieved evidence for one forecasting task and elicits one likelihood per item. The Bayesian Linguistic Forecaster maintains a structured belief state during iterative search and aggregates calibrated forecasts across trials \citep{murphy2026agentic}; LEAP keeps evidence collection and prediction separate, then computes a posterior after the evidence set is fixed. This design is motivated by known weaknesses of holistic synthesis, including long-context underuse \citep{liu2024lost}, unsupported factual claims in long outputs \citep{min2023factscore}, and hallucination relative to supplied sources \citep{niu2024ragtruth,ji2023survey,belem2025single}, as well as probabilistic forecasting principles that favor calibrated distributions under proper scoring rules \citep{guo2017calibration,brier1950verification,gneiting2007strictly}. Because LEAP only requires a task, a fixed evidence set, and an output type, it can plug into different forecasting agents, browsing agents, or agent CLI frameworks. The output type selects the local likelihood schema and posterior family, but the collection interface stays fixed. This interface supports continuous, single-choice, and multi-choice targets across different upstream systems. The LLM interprets each evidence item locally, while an explicit probabilistic model performs the final combination, yielding forecasts that are more auditable than a single free-text rationale.

\vfill\newpage

\section{Problem Formulation}
\label{sec:problem}

\subsection{Task and Two-Stage Decomposition}
\label{sec:two-stage}

A forecasting task is specified by a tuple
\begin{equation}
T = (q,\ t_{\mathrm{freeze}},\ \tau,\ \mathcal{O}),
\end{equation}
where $q$ is the question; $t_{\mathrm{freeze}}$ is the cutoff time, so usable information must be timestamped no later than this time; $\tau$ is the output type, either continuous, single-choice, or multi-choice; and $\mathcal{O}$ is an optional set of candidate options for discrete outputs. We decompose evidence-grounded LLM forecasting into two stages. A \emph{collection} stage uses an agent loop to gather evidence items
\begin{equation}
\mathcal{E} = \{e_1, \ldots, e_n\},
\end{equation}
where each $e_i$ is a text passage with a timestamp at or before $t_{\mathrm{freeze}}$. A \emph{prediction} stage then maps $(T, \mathcal{E})$ to a forecast $f$. The prediction stage may read $\mathcal{E}$ but performs no additional retrieval.

\subsection{Forecast Targets}
\label{sec:forecast-targets}

The output type $\tau$ determines both the latent quantity and the form of the forecast $f$. We consider three types: \emph{continuous} targets for numeric quantities, \emph{single-choice} targets where exactly one of $K$ options is correct, and \emph{multi-choice} targets where any subset of $K$ options may be correct.

\paragraph{Continuous.}
The latent target $\theta \in \mathbb{R}$ is a real-valued quantity, such as a closing price or a published economic indicator. The forecast reports values at fixed quantile levels, such as the median and the $10$th and $90$th percentiles, capturing both central tendency and uncertainty.

\paragraph{Single-choice.}
The latent target $\theta \in \{1, \ldots, K\}$ is exactly one of the $K$ candidate options in $\mathcal{O}$. The forecast is a probability distribution over these options: a vector $f \in [0, 1]^K$ whose entries sum to one, with $f_k$ giving the probability that option $k$ is the correct one.

\paragraph{Multi-choice.}
The latent target $\theta \in \{0, 1\}^K$ is a vector of yes/no labels, one per option in $\mathcal{O}$. The forecast $f \in [0, 1]^K$ assigns one probability to each option, with $f_k$ being the probability that option $k$ is true. Because options are evaluated independently rather than as alternatives, the entries of $f$ need not sum to one.

\subsection{Monolithic Baseline and Evaluation Protocol}
\label{sec:protocol}
The comparison baseline is Monolithic Prediction. It uses the same LLM and the same $\mathcal{E}$ as our method, reads $(T, \mathcal{E})$ in a single prompt, and emits a forecast $f$ in the form specified by $\tau$ in one pass.

Operationally, both methods share the same collection stage and are applied to the same $(T, \mathcal{E})$; they differ only in how that evidence is consumed during prediction. For continuous targets, \emph{Monolithic} would normally return a single number; we additionally ask it to provide an uncertainty range around that number so continuous predictions can be compared under the same scoring setup.

For each task $T$, the collection stage is run once to produce a single $\mathcal{E}$, and both methods are applied to the same $(T, \mathcal{E})$. Score differences can therefore be attributed to how evidence is turned into a forecast, not to retrieval differences or search budget.

\section{Method}
\label{sec:method}

\begin{figure*}[t]
\centering
\includegraphics[width=.98\textwidth]{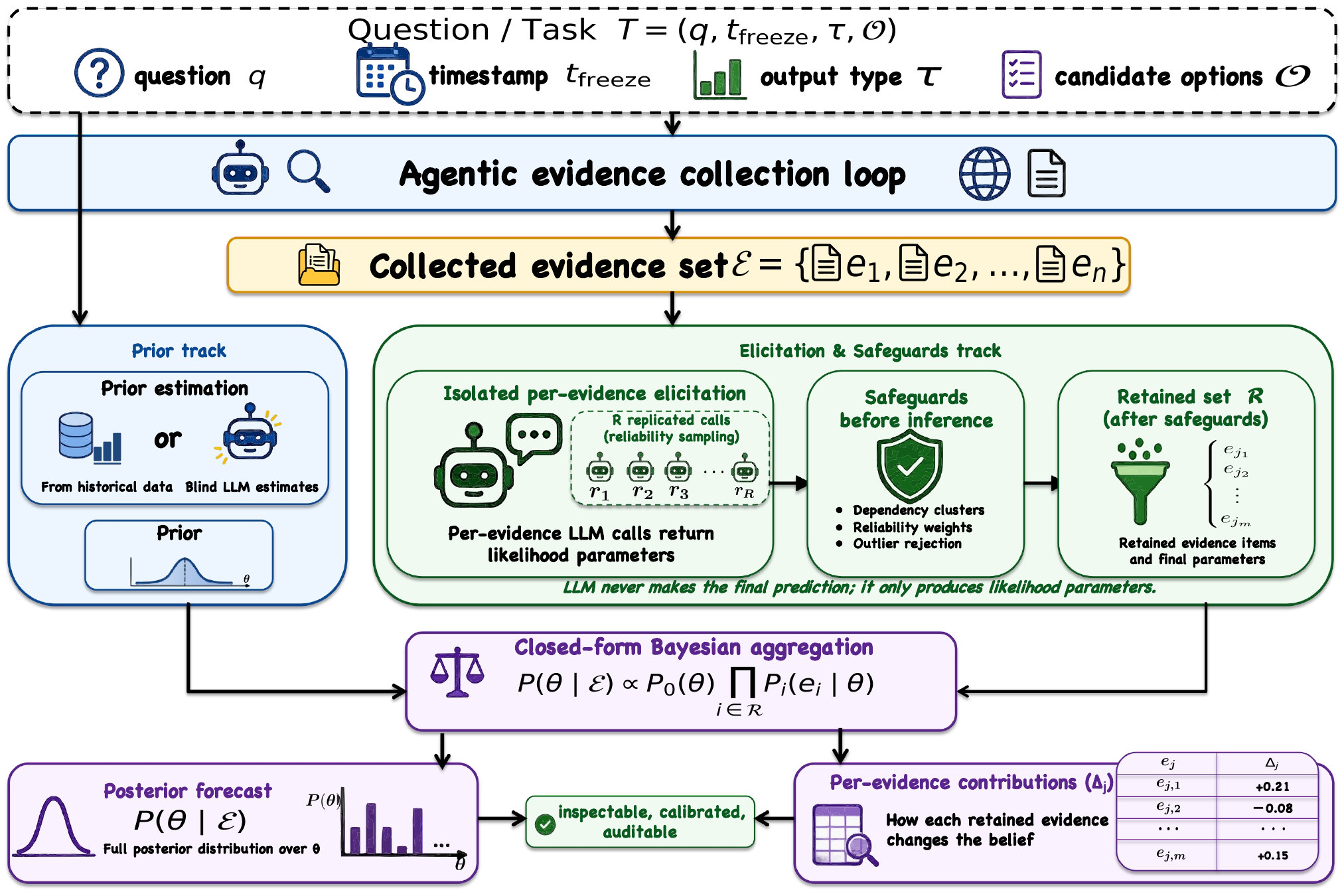}
\caption{Overview of LEAP. After evidence collection, isolated LLM calls estimate a prior and one likelihood per evidence item. LEAP applies safeguards, performs a closed-form Bayesian update, and reports the posterior forecast with leave-one-out contributions $\Delta_j$.}
\label{fig:pipeline}
\end{figure*}

LEAP instantiates the prediction stage of Section~\ref{sec:two-stage}: it consumes $(T,\mathcal{E})$ and returns a forecast $f$ in the form specified by $\tau$. Its key design choice is not to ask the LLM for $f$. Instead, prediction is divided into local parameter elicitation, in which the LLM analyzes one evidence item at a time and estimates parameters for an explicit probabilistic model, and deterministic aggregation, in which the probabilistic model combines these parameters with a prior through a closed-form update to produce the final prediction. The LLM thus supplies local evidence interpretations, while the probabilistic model makes the final judgment after all evidence items have been analyzed. Figure~\ref{fig:pipeline} illustrates the complete procedure, including the upstream collection loop that constructs $\mathcal{E}$.

\subsection{Probabilistic Model}
\label{sec:prob-model}

LEAP starts from a Bayesian probability model on the prediction target $\theta$, the unknown answer variable, and the evidence in $\mathcal{E}$:
\begin{equation}
\theta \sim P_0(\theta), \qquad e_i \mid \theta \;\overset{\text{indep}}{\sim}\; P_i(e_i \mid \theta),
\label{eq:gen-model}
\end{equation}
for $i = 1, \ldots, n$. The prior $P_0$ encodes a base rate for $\theta$ before any evidence is examined. Each conditional $P_i$ describes what the appearance of $e_i$ implies about $\theta$, under the working conditional independence assumption $e_i \perp e_j \mid \theta$ for $i \neq j$. Because duplicated or common-source evidence can violate this, LEAP later clusters dependent evidence before aggregation. For any retained subset $\mathcal{R} \subseteq \{1, \ldots, n\}$, the unweighted posterior follows Bayes' rule:
\begin{equation}
P(\theta \mid \mathcal{E}) \;\propto\; P_0(\theta) \prod_{i \in \mathcal{R}} P_i(e_i \mid \theta).
\label{eq:posterior}
\end{equation}
We instantiate $P_0$ and $\{P_i\}$ as conjugate pairs so Equation~\eqref{eq:posterior} is closed form. Our experiments use a tempered version with role weights, reducing to Equation~\eqref{eq:posterior} at unit temperature and weights. The three output types of Section~\ref{sec:forecast-targets} use a Gaussian conjugate pair for continuous targets, a categorical pair with multinomial likelihood for single-choice targets, and independent Bernoulli pairs for multi-choice targets; closed-form updates are in Appendix~\ref{app:posteriors}.

\subsection{Parameter Elicitation}
\label{sec:elicit}

The LLM estimates parameters of $P_0$ and each $P_i$; it does not output a forecast over $\theta$. This is a constrained use of the model's reasoning ability: modern LLMs are trained and evaluated on analytical and mathematical tasks, so LEAP uses them to estimate local likelihood signals while leaving the posterior update to the probabilistic model. Each evidence-level call analyzes $T$ with one evidence item and returns a structured likelihood schema. The call may see that item, but not the full evidence set, other evidence items, accumulated evidence, or any partial posterior. It therefore returns inputs to likelihood factors, not an answer or a final probability. Elicitation has two tasks: estimating prior parameters before any evidence is consulted, and estimating likelihood parameters for each $e_i$ from that evidence item alone.

\paragraph{Prior parameters.}
The parameters of $P_0$ come from the most informative source available. When the target has a recent and reliable numerical history, such as a market price or macroeconomic index, the prior is computed from that history. Otherwise, a single LLM call with no evidence visible returns prior parameters from background knowledge alone. For discrete tasks that lack even a reliable background base rate, we use an unbiased and conservative prior. The prompt forbids conditioning on specific evidence, since the prior should record a base rate before evidence is examined; letting the prior call read evidence would count that evidence twice in Equation~\eqref{eq:posterior}.

\paragraph{Evidence likelihood parameters.}

Each evidence item $e_i$ is sent to the LLM in isolation: the model sees $T$ and exactly one evidence item, never other sources, accumulated evidence, or partial posteriors. The prompt does not ask for a prediction; it elicits how $e_i$ bears on each candidate outcome. The output is a structured qualitative schema for $P_i$, plus a \emph{dependency key} identifying the source family of $e_i$.

For continuous targets, the LLM extracts a target-scale observation from $e_i$: an explicit reported value when present, or a qualitative assessment otherwise. This observation becomes the likelihood mean $\mu_i$. The likelihood standard deviation $\sigma_i$ reflects evidence strength, determined by the qualitative label and sampling agreement. Appendix~\ref{app:mappings} details the schema and strength calibration.

For single-choice targets, the LLM returns qualitative support labels for each option $k$, which determine non-normalized likelihoods $L_i(k)=P(e_i\mid\theta=k)$. For multi-choice targets, it returns support and opposition labels that determine log likelihood ratios. These mappings follow standard elicitation practice for converting qualitative assessments into likelihood parameters. Appendix~\ref{app:mappings} gives the complete schema.

\subsection{Deterministic Aggregation}
\label{sec:aggregation}

Given the elicited prior and likelihood parameters and the retained evidence set $\mathcal{R}$, aggregation is deterministic and uses no additional LLM call: it evaluates the posterior over the retained representatives, reads out the forecast in the form required by $\tau$, and decomposes the update into evidence contributions.

\paragraph{Posterior update.}
The tempered posterior update is evaluated over $\mathcal{R}$ using the closed-form conjugate update for the chosen family. The forecast $f$ is a direct readout of the resulting distribution: quantiles for continuous targets, normalised option probabilities for single-choice targets, and Bernoulli marginals for multi-choice targets.

\paragraph{Per-evidence contribution.}
For each retained item $j \in \mathcal{R}$, we recompute the update with $j$ excluded and report the resulting forecast change as a leave-one-out contribution $\Delta_j$. Because the computation is deterministic and closed form, $\Delta_j$ is reproducible by running the update on $\mathcal{R} \setminus \{j\}$. Appendix~\ref{app:posteriors} gives the concrete form for each $\tau$.

\subsection{Safeguard Components}
\label{sec:safeguards}

\paragraph{Dependency clustering.}
The conditional independence assumption in Equation~\eqref{eq:gen-model} can fail when evidence items share a source, e.g., multiple pages quoting one report. LEAP uses the dependency key returned with each likelihood to group common-source evidence and retain one representative per group; only the retained set $\mathcal{R}$ enters the posterior update. Appendix~\ref{app:safeguards} describes how equivalent keys are canonicalised.

\paragraph{Reliability sampling.}
A single local likelihood query may be noisy. LEAP can repeat the same query for an evidence item $R$ times and use agreement across draws as a reliability adjustment. These draws are not a forecast ensemble: each draw still sees only $(T,e_i)$ and returns the same local schema. Agreement only shrinks or preserves the local likelihood magnitude, and the final forecast is still produced once by the deterministic aggregation step.

\paragraph{Outlier rejection.}
For continuous targets with a data-derived prior, the prior serves as a numerical anchor for implausible elicited observations: if an evidence likelihood mean exceeds four prior standard deviations from the prior mean, the item is excluded from $\mathcal{R}$ as a likely elicitation error (e.g., a unit mismatch). This rule is not applied when the prior is itself elicited without historical data.

\section{Experiments}
\label{sec:experiments}

\subsection{Setup}
\label{sec:setup}

\paragraph{Benchmark.}
The benchmark draws forecasting and information-seeking tasks from FutureX \citep{zeng2025futurex}, GAIA \citep{mialon2024gaia}, and BrowseComp \citep{wei2025browsecomp}, and converts each task into the tuple of Section~\ref{sec:two-stage}. The output type $\tau$ follows the ground truth answer type, and ranking resolutions are excluded. Appendix~\ref{app:experiments} reports counts, conversion rules, $t_{\mathrm{freeze}}$ assignment, and the temporal leakage audit.

\paragraph{Evaluation settings.}
We evaluate LEAP in two settings, both using a single evidence set $\mathcal{E}$ shared by \emph{Monolithic} and LEAP for each task. To avoid evidence leakage, retrieval is restricted to material available no later than the task timestamp, and the base models are chosen so that their knowledge cutoffs precede the evaluated timestamps. The first setting uses our ReAct-style agent loop across five base models: DeepSeek-V3.2, Gemini-3.1-Flash-Lite, Claude-Haiku-4.5, GPT-5.4-mini, and Grok-4.20-Fast. The second applies LEAP as a probability skill on unmodified traces from four external agent CLI frameworks: DeerFlow, Hermes, OpenClaw, and MiroFlow.
\paragraph{Metrics.}
All metrics are computed on our constructed benchmark. The FutureX composite score \citep{zeng2025futurex} is the headline metric across all output types. Accuracy, Brier score \citep{brier1950verification}, and Spherical score \citep{gneiting2007strictly} are computed on discrete tasks; NCRPS, a length-normalised variant of the continuous ranked probability score, is computed on continuous tasks \citep{gneiting2007strictly}. Calibration diagnostics, including expected calibration error and overconfidence, are reported in Section~\ref{sec:analysis}, with formal definitions in Appendix~\ref{app:experiments}.

\begin{table*}[t]
\centering
\small
\renewcommand{\arraystretch}{1.15}
\setlength{\tabcolsep}{4pt}
\begin{tabularx}{\textwidth}{@{}l l Y Y Y Y Y@{}}
\toprule
\textbf{Model} & \textbf{Method} & \textbf{FutureX}$\uparrow$ & \textbf{Accuracy}$\uparrow$ & \textbf{Brier}$\downarrow$ & \textbf{Spherical}$\uparrow$ & \textbf{NCRPS}$\downarrow$ \\
\midrule
\multirow{2}{*}{DeepSeek-V3.2}
  & \emph{Monolithic} & 0.4895 & 0.6237 & \textbf{0.2308} & 0.5581 & 0.2594 \\
  & LEAP   & \textbf{0.6701} & \textbf{0.6407} & 0.2374 & \textbf{0.7089} & \textbf{0.2578} \\
\midrule
\multirow{2}{*}{Gemini-3.1-Flash-Lite}
  & \emph{Monolithic} & 0.6004 & 0.6271 & \textbf{0.2334} & 0.6693 & 0.4916 \\
  & LEAP   & \textbf{0.7069} & \textbf{0.6542} & 0.2336 & \textbf{0.7223} & \textbf{0.1419} \\
\midrule
\multirow{2}{*}{Claude-Haiku-4.5}
  & \emph{Monolithic} & 0.6211 & 0.5933 & 0.2638 & 0.6726 & 0.2432 \\
  & LEAP   & \textbf{0.6569} & \textbf{0.6102} & \textbf{0.2522} & \textbf{0.6975} & \textbf{0.1605} \\
\midrule
\multirow{2}{*}{GPT-5.4-mini}
  & \emph{Monolithic} & 0.6456 & 0.6473 & 0.2741 & 0.6857 & 0.5413 \\
  & LEAP   & \textbf{0.7222} & \textbf{0.6873} & \textbf{0.2100} & \textbf{0.7484} & \textbf{0.0200} \\
\midrule
\multirow{2}{*}{Grok-4.20-Fast}
  & \emph{Monolithic} & 0.6533 & 0.6314 & 0.2971 & 0.6750 & 0.1282 \\
  & LEAP   & \textbf{0.7503} & \textbf{0.7133} & \textbf{0.1916} & \textbf{0.7571} & \textbf{0.1174} \\
\bottomrule
\end{tabularx}
\caption{Main comparison of \emph{Monolithic} and LEAP across five base models under the evaluation protocol of Section~\ref{sec:problem}, with our ReAct-style agent loop fixed across rows. Arrows indicate the better direction. The better value per (model, metric) is in bold.}
\label{tab:main}
\end{table*}

\begin{table*}[t]
\centering
\small
\renewcommand{\arraystretch}{1.15}
\setlength{\tabcolsep}{4pt}
\begin{tabularx}{\textwidth}{@{}l l Y Y Y Y Y@{}}
\toprule
\textbf{Framework} & \textbf{Method} & \textbf{FutureX}$\uparrow$ & \textbf{Accuracy}$\uparrow$ & \textbf{Brier}$\downarrow$ & \textbf{Spherical}$\uparrow$ & \textbf{NCRPS}$\downarrow$ \\
\midrule
\multirow{2}{*}{DeerFlow}
  & \emph{Monolithic} & 0.4135 & 0.3500 & 0.5757 & 0.3790 & 1.0000 \\
  & LEAP   & \textbf{0.4832} & 0.3500 & \textbf{0.4360} & \textbf{0.4902} & \textbf{0.5024} \\
\midrule
\multirow{2}{*}{Hermes}
  & \emph{Monolithic} & 0.4238 & 0.3810 & 0.4925 & 0.4678 & 0.8661 \\
  & LEAP   & \textbf{0.5619} & \textbf{0.4762} & \textbf{0.2552} & \textbf{0.6723} & \textbf{0.6721} \\
\midrule
\multirow{2}{*}{OpenClaw}
  & \emph{Monolithic} & 0.3472 & 0.1667 & 0.5296 & 0.3974 & 0.5142 \\
  & LEAP   & \textbf{0.4396} & \textbf{0.2222} & \textbf{0.3674} & \textbf{0.5413} & \textbf{0.4808} \\
\midrule
\multirow{2}{*}{MiroFlow}
  & \emph{Monolithic} & 0.5630 & 0.5185 & 0.3683 & 0.5912 & 0.7611 \\
  & LEAP   & \textbf{0.6545} & \textbf{0.5556} & \textbf{0.2393} & \textbf{0.7016} & \textbf{0.7593} \\
\midrule
\multirow{2}{*}{Overall}
  & \emph{Monolithic} & 0.4491 & 0.3721 & 0.4806 & 0.4712 & 0.7598 \\
  & LEAP   & \textbf{0.5471} & \textbf{0.4186} & \textbf{0.3157} & \textbf{0.6117} & \textbf{0.6312} \\
\bottomrule
\end{tabularx}
\caption{\emph{Monolithic} vs LEAP applied as a downstream probability skill on evidence collected by four external agent CLI frameworks. Arrows indicate the better direction. The better value per (framework, metric) is in bold. The Overall row is the macro-average across the four frameworks.}
\label{tab:agentos}
\end{table*}

\subsection{Main Results}
\label{sec:main-results}

\paragraph{Comparison across models on our Agent Loop.}
Table~\ref{tab:main} compares \emph{Monolithic} and LEAP with our ReAct-style agent loop fixed and the base model varied. For every base model, LEAP improves FutureX, Spherical score, and accuracy. Absolute gains range from 3.6 to 18.1 points on FutureX and from 2.5 to 15.1 points on Spherical. NCRPS also improves for every base model, with gains above 50 points on GPT-5.4-mini and 30 points on Gemini-3.1-Flash-Lite. Brier is less uniform: LEAP improves it on three of five base models, while the other two are close or essentially tied. Section~\ref{sec:analysis} interprets this with calibration diagnostics.

\paragraph{Robustness across external agent CLI frameworks.}
Table~\ref{tab:agentos} reports the external agent CLI setting. The pattern from our loop largely carries over: across four frameworks, LEAP improves most reported metrics over \emph{Monolithic}, with FutureX gains of 7.0 to 13.8 points and Brier reductions of 12.9 to 23.7 points. The macro-average row improves on all five metrics, with gains of 9.8 points on FutureX, 4.7 on accuracy, 16.5 on Brier, 14.1 on Spherical, and 12.9 on NCRPS. This suggests that LEAP can improve existing forecasting systems as a plug-in module without modifying their collection pipelines.

\subsection{Analysis}
\label{sec:analysis}

We complement the main results with diagnostics on calibration, forecast horizons, and LEAP components, plus case studies showing evidence-level contributions. Appendix~\ref{app:experiments} describes the diagnostic subset and base model.

\begin{figure}[t]
\centering
\includegraphics[width=\columnwidth]{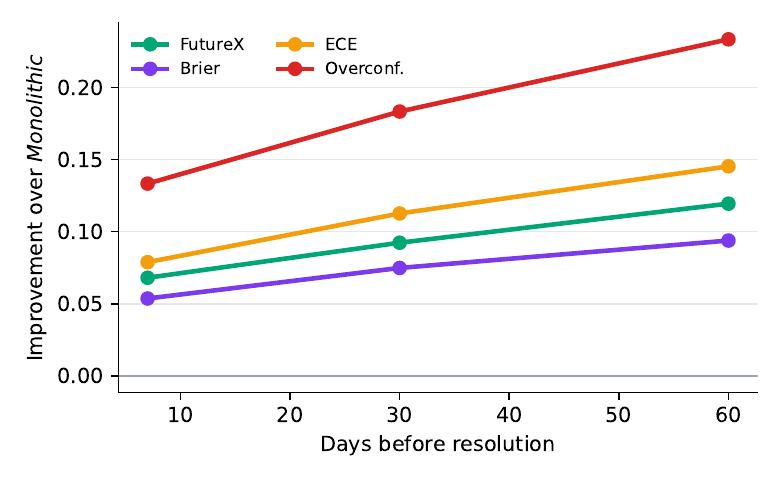}
\caption{Absolute improvement of LEAP over \emph{Monolithic} on the diagnostic subset as a function of the forecast horizon (days between evidence collection and ground truth resolution). All four reported metrics move further in LEAP's favour as the horizon lengthens.}
\label{fig:leadtime}
\end{figure}

\paragraph{Calibration.}
Table~\ref{tab:calibration} explains why Brier is the least uniform metric in Table~\ref{tab:main}. Following standard calibration evaluation \citep{guo2017calibration}, ECE measures the gap between top-option confidence and empirical accuracy; Adaptive ECE uses data-dependent bins, and overconfidence averages the positive part of this gap. Although LEAP is close to or slightly behind \emph{Monolithic} on Brier for DeepSeek-V3.2 and Gemini-3.1-Flash-Lite, it roughly halves both ECE variants (0.184 to 0.088, and 0.177 to 0.091) and reduces overconfidence by more than half (0.317 to 0.150). This indicates a calibration difference rather than a loss in predictive quality: \emph{Monolithic} concentrates probability on one option, lowering Brier when it is right but paying with sharp overconfidence on the rest. LEAP spreads probability more conservatively and keeps confidence aligned with accuracy.

\paragraph{Robustness across forecast horizons.}
Forecasting becomes harder as the horizon between evidence collection and ground truth resolution lengthens. We split the subset into horizons of 7, 30, and 60 days and plot LEAP's absolute improvement over \emph{Monolithic} in Figure~\ref{fig:leadtime}. All four curves rise monotonically: FutureX gain increases from 6.8 to 11.9 points, and the overconfidence gap widens from 13.3 to 23.3 points. As evidence becomes more indirect, \emph{Monolithic} continues to commit firmly to a single option, while LEAP recognises weaker support in $\mathcal{E}$ and produces wider posteriors—degrading by becoming conservative rather than confidently wrong.

\paragraph{Component ablation.}
Table~\ref{tab:ablation} reports the results of removing the prior, dependency clustering, and reliability sampling one at a time.

\begin{table}[t]
\centering
\footnotesize
\renewcommand{\arraystretch}{1.15}
\setlength{\tabcolsep}{4pt}
\begin{tabular}{@{}l r r r@{}}
\toprule
\textbf{Method} & \textbf{FutureX}$\uparrow$ & \textbf{Brier}$\downarrow$ & \textbf{ECE}$\downarrow$ \\
\midrule
\emph{Monolithic} & 0.6512 & 0.2689 & 0.1840 \\
\midrule
LEAP w/o prior            & 0.6427 & 0.2808 & 0.2134 \\
LEAP w/o dep.\ clustering & 0.7089 & 0.2299 & 0.1578 \\
LEAP w/o reliab.\ sampling & 0.7136 & 0.2215 & 0.1219 \\
Full LEAP                  & \textbf{0.7284} & \textbf{0.2057} & \textbf{0.0876} \\
\bottomrule
\end{tabular}
\caption{Component ablation on the diagnostic subset. Each row removes one component from Full LEAP; \emph{Monolithic} is reported for reference. Best values are in bold.}
\label{tab:ablation}
\end{table}

\emph{Prior.} Removing the prior drops FutureX below \emph{Monolithic} (0.643 versus 0.651) and raises ECE to 0.213, making it the most important component on this subset: structured likelihood aggregation needs a stable starting point when local evidence is sparse or noisy.

\emph{Dependency clustering.} Removing dependency clustering lowers FutureX by 1.9 points and raises ECE from 0.088 to 0.158, consistent with overcounting common source evidence and making the posterior sharper than the evidence warrants.

\emph{Reliability sampling.} Removing reliability sampling has a smaller but consistent effect, lowering FutureX to 0.714 and raising ECE to 0.122; repeated local checks mainly protect against noisy likelihood estimates rather than supplying the main gain. All variants except the one without the prior remain above \emph{Monolithic} on FutureX, so structured aggregation supplies most of the gain while these safeguards add robustness.

\begin{table*}[t]
\centering
\small
\renewcommand{\arraystretch}{1.12}
\setlength{\tabcolsep}{4pt}
\begin{tabularx}{\textwidth}{@{}X r c r r r@{}}
\toprule
\textbf{Method} & \textbf{Tokens/task} & \textbf{p50 / p95 (s)} & \textbf{FX $\uparrow$} & \textbf{Brier $\downarrow$} & \textbf{ECE $\downarrow$} \\
\midrule
\emph{Monolithic} & $5{,}733$ & $6.2 / 12.1$ & $0.6512$ & $0.2689$ & $0.1840$ \\
\emph{Monolithic} + same prior, one sample & $6{,}521$ & $9.1 / 17.0$ & $0.6742$ & $0.2510$ & $0.1508$ \\
LEAP without reliability sampling & $5{,}869$ & $7.3 / 16.5$ & $0.7136$ & $0.2215$ & $0.1219$ \\
\emph{Monolithic} + same prior, two-sample mean & $11{,}753$ & $9.8 / 20.4$ & $0.6736$ & $0.2497$ & $0.1471$ \\
LEAP & $11{,}508$ & $10.4 / 27.8$ & $\mathbf{0.7284}$ & $\mathbf{0.2057}$ & $\mathbf{0.0876}$ \\
\bottomrule
\end{tabularx}
\caption{Prediction-time budget and latency under the diagnostic setup in Appendix~\ref{app:diagnostic-subset}. The prior-matched \emph{Monolithic} baselines control for prior access, and the two-sample mean controls for LEAP's token budget.}
\label{tab:controlled-budget}
\end{table*}

\begin{table}[t]
\centering
\small
\renewcommand{\arraystretch}{1.12}
\setlength{\tabcolsep}{4pt}
\begin{tabular*}{\columnwidth}{@{\extracolsep{\fill}}l r r r@{}}
\toprule
\textbf{Method} & \textbf{FX $\uparrow$} & \textbf{Brier $\downarrow$} & \textbf{ECE $\downarrow$} \\
\midrule
\multicolumn{4}{@{}l}{\emph{Aggregation}} \\
\emph{Monolithic} & $0.6512$ & $0.2689$ & $0.1840$ \\
Linear opinion pool & $0.6808$ & $0.2389$ & $0.1117$ \\
LEAP ($\eta=1.0$) & $\mathbf{0.7284}$ & $\mathbf{0.2057}$ & $\mathbf{0.0876}$ \\
\midrule
\multicolumn{4}{@{}l}{\emph{Likelihood strength for LEAP}} \\
$\eta=0.5$ & $0.7181$ & $0.2166$ & $0.1012$ \\
$\eta=1.0$ & $\mathbf{0.7284}$ & $\mathbf{0.2057}$ & $\mathbf{0.0876}$ \\
$\eta=2.0$ & $0.7210$ & $0.2139$ & $0.1094$ \\
\bottomrule
\end{tabular*}
\caption{Aggregation baseline and sensitivity to the global likelihood strength $\eta$, which scales the contribution of elicited likelihoods in the posterior update. The default is $\eta=1.0$.}
\label{tab:aggregation-sensitivity}
\end{table}

\begin{table}[!t]
\centering
\small
\renewcommand{\arraystretch}{1.15}
\setlength{\tabcolsep}{5pt}
\begin{tabular*}{\columnwidth}{@{\extracolsep{\fill}} l r r @{}}
\toprule
\textbf{Metric} & \textbf{\emph{Monolithic}} & \textbf{LEAP} \\
\midrule
FutureX$\uparrow$           & 0.6512 & \textbf{0.7284} \\
Brier$\downarrow$           & 0.2689 & \textbf{0.2057} \\
ECE$\downarrow$             & 0.1840 & \textbf{0.0876} \\
Adaptive ECE$\downarrow$    & 0.1765 & \textbf{0.0912} \\
Overconfidence$\downarrow$  & 0.3167 & \textbf{0.1500} \\
\bottomrule
\end{tabular*}
\caption{Calibration diagnostics on the diagnostic subset. Arrows indicate the better direction; bold marks the better value.}
\label{tab:calibration}
\end{table}

\paragraph{Controlled baselines and inference efficiency.}
We also run controlled experiments to measure LEAP's token cost and inference efficiency. Tables~\ref{tab:controlled-budget} and~\ref{tab:aggregation-sensitivity} show that LEAP retains its advantage when \emph{Monolithic} receives the same prior or a comparable token budget, and that it performs better than linear opinion pooling. Its median latency is close to that of the two-sample ensemble, while its p95 latency is higher because completion time depends on the slowest evidence-level calls. Performance changes little across the tested likelihood strengths. Appendix~\ref{app:rebuttal-controls} reports alternative evidence grouping strategies and source-wise results.

\paragraph{Details and extended experiments.}
Appendix~\ref{app:experiments} reports prompts, the ReAct-style agent loop, the probability skill for external traces, additional sensitivity analyses, and case studies illustrating typical LEAP successes, failure modes, and how $\Delta_j$ makes forecasts auditable at the evidence-item level.

\section{Conclusion}
\label{sec:conclusion}

This paper studied how LLM-based forecasting systems turn a fixed set of gathered evidence into a final prediction. Building on work that uses LLM judgments within probabilistic models, we proposed LEAP for this setting. LEAP leaves evidence collection unchanged, interprets each evidence item locally, and combines the resulting likelihoods with a prior through deterministic probabilistic aggregation.

Under a protocol that holds evidence fixed across methods, LEAP improves prediction accuracy and probabilistic quality in most settings and metrics across five base models and four external agent CLI frameworks. Calibration diagnostics show that LEAP roughly halves expected calibration error and overconfidence, with gains widening as the forecast horizon lengthens. Because inference is closed form, evidence contributions are reproducible by removing one item and rerunning the posterior update, yielding an audit signal not available from a free-text LLM rationale.

\section*{Limitations}

The scope of LEAP is the prediction step after evidence has already been collected. It changes how a fixed evidence set is converted into a forecast, but it does not improve the upstream collection stage itself. If the collected evidence is sparse, outdated, or only weakly relevant, LEAP will usually express that uncertainty in a wider posterior rather than recover information that was never retrieved.

Our evaluation is also bounded by the benchmark and model families studied here. We evaluate English forecasting and agentic information-seeking tasks with a fixed collection protocol and a limited set of base models. Extending the evaluation to other languages, specialised scientific domains, longer forecast horizons, and substantially different agent designs is left to future work. LEAP also uses more inference calls than \emph{Monolithic}, since evidence items are interpreted separately and may be sampled more than once for local verification. This cost is the main practical tradeoff for the added calibration and auditability.

\section*{Ethical Considerations}

This work studies probabilistic forecasting systems that use LLMs. Their outputs should be understood as decision support rather than authoritative predictions. Forecasts about financial, political, public health, or similarly consequential events may affect downstream decisions if deployed in practice. Such systems should therefore be used with human oversight, clear uncertainty communication, and review by domain experts, especially in settings where incorrect forecasts could cause material harm.

LEAP is designed to make the prediction step more inspectable by exposing retained evidence items and their contributions, but this audit trail does not eliminate the risks inherited from the upstream evidence collection stage or from the underlying LLM. Retrieved sources may be incomplete, biased, outdated, or unevenly representative, and the LLM's local interpretations may still reflect model biases or factual errors. The method should not be presented as removing the need for source verification or independent expert judgment.

Our experiments use benchmark tasks and collected evidence snapshots rather than private user data. We use existing benchmark artifacts only for research evaluation, and any converted derivatives are intended to remain research-only artifacts under the original access conditions, including the gated-access constraints attached to GAIA-derived records. More broadly, deployments that collect external evidence should respect data access policies, copyright restrictions, and privacy expectations of the sources they use. The same forecasting capabilities could also be misused to support manipulation, speculation, or strategic targeting; these risks are not specific to LEAP, but they are relevant to any system that improves the scalability or apparent credibility of automated forecasting.

\bibliography{custom}

@inproceedings{zou2022forecasting,
  author    = {Andy Zou and Tristan Xiao and Ryan Jia and Joe Kwon and Mantas Mazeika and Richard Li and Dawn Song and Jacob Steinhardt and Owain Evans and Dan Hendrycks},
  title     = {Forecasting Future World Events with Neural Networks},
  booktitle = {Advances in Neural Information Processing Systems},
  volume    = {35},
  pages     = {27293--27305},
  publisher = {Curran Associates, Inc.},
  year      = {2022},
  doi       = {10.52202/068431-1979},
  url       = {https://proceedings.neurips.cc/paper_files/paper/2022/hash/aec870a6772336c15dac992c16f2e7c9-Abstract-Datasets_and_Benchmarks.html}
}

@inproceedings{yan2024autocastpp,
  author    = {Qi Yan and Raihan Seraj and Jiawei He and Lili Meng and Tristan Sylvain},
  title     = {{AutoCast++}: Enhancing World Event Prediction with Zero-shot Ranking-based Context Retrieval},
  booktitle = {The Twelfth International Conference on Learning Representations},
  pages     = {34523--34539},
  publisher = {OpenReview.net},
  year      = {2024},
  url       = {https://proceedings.iclr.cc/paper_files/paper/2024/hash/93f01c8d9b355d7bbe3f353b44ccde66-Abstract-Conference.html}
}

@inproceedings{karger2025forecastbench,
  author    = {Ezra Karger and Houtan Bastani and Chen Yueh{-}Han and Zachary Jacobs and Danny Halawi and Fred Zhang and Philip E. Tetlock},
  title     = {{ForecastBench}: A Dynamic Benchmark of {AI} Forecasting Capabilities},
  booktitle = {The Thirteenth International Conference on Learning Representations},
  pages     = {93943--93980},
  publisher = {OpenReview.net},
  year      = {2025},
  url       = {https://proceedings.iclr.cc/paper_files/paper/2025/hash/ea74e45a229dac70b5b63b28d8934db6-Abstract-Conference.html}
}

@inproceedings{zeng2025futurex,
  author    = {Zhiyuan Zeng and Jiashuo Liu and Siyuan Chen and Tianci He and Yali Liao and Yixiao Tian and Jinpeng Wang and Zaiyuan Wang and Yang Yang and Lingyue Yin and Mingren Yin and Zhenwei Zhu and Tianle Cai and Zehui Chen and Jiecao Chen and Yantao Du and Xiang Gao and Jiacheng Guo and Liang Hu and Jianpeng Jiao and Xiangsheng Li and Jingkai Liu and Shuang Ni and Zhoufutu Wen and Ge Zhang and Kaiyuan Zhang and Xin Zhou and Wenhao Huang and Jose H. Blanchet and Xipeng Qiu and Mengdi Wang},
  title     = {{FutureX}: An Advanced Live Benchmark for {LLM} Agents in Future Prediction},
  booktitle = {The Fourteenth International Conference on Learning Representations},
  pages     = {19522--19563},
  publisher = {OpenReview.net},
  year      = {2026},
  url       = {https://proceedings.iclr.cc/paper_files/paper/2026/hash/21315abf210cef71317e887e9cda78b3-Abstract-Conference.html}
}

@inproceedings{halawi2024approaching,
  author    = {Danny Halawi and Fred Zhang and Chen Yueh{-}Han and Jacob Steinhardt},
  title     = {Approaching Human-Level Forecasting with Language Models},
  booktitle = {Advances in Neural Information Processing Systems},
  volume    = {37},
  pages     = {50426--50468},
  publisher = {Curran Associates, Inc.},
  year      = {2024},
  doi       = {10.52202/079017-1598},
  url       = {https://proceedings.neurips.cc/paper_files/paper/2024/hash/5a5acfd0876c940d81619c1dc60e7748-Abstract-Conference.html}
}

@article{alur2025aia,
  author  = {Rohan Alur and Bradly C. Stadie and Daniel Kang and Ryan Chen and Matt McManus and Michael Rickert and Tyler Lee and Michael Federici and Richard Zhu and Dennis Fogerty and Hayley Williamson and Nina Lozinski and Aaron Linsky and Jasjeet S. Sekhon},
  title   = {{AIA} Forecaster: Technical Report},
  journal = {CoRR},
  volume  = {abs/2511.07678},
  year    = {2025},
  doi     = {10.48550/arXiv.2511.07678},
  url     = {https://doi.org/10.48550/arXiv.2511.07678}
}

@article{schoenegger2024wisdom,
  author  = {Philipp Schoenegger and Indre Tuminauskaite and Peter S. Park and Rafael Valdece Sousa Bastos and Philip E. Tetlock},
  title   = {Wisdom of the silicon crowd: {LLM} ensemble prediction capabilities rival human crowd accuracy},
  journal = {Science Advances},
  volume  = {10},
  number  = {45},
  pages   = {eadp1528},
  year    = {2024},
  doi     = {10.1126/sciadv.adp1528},
  url     = {https://doi.org/10.1126/sciadv.adp1528}
}

@inproceedings{feng2025bird,
  author    = {Yu Feng and Ben Zhou and Weidong Lin and Dan Roth},
  title     = {{BIRD}: A Trustworthy Bayesian Inference Framework for Large Language Models},
  booktitle = {The Thirteenth International Conference on Learning Representations},
  pages     = {8961--8989},
  publisher = {OpenReview.net},
  year      = {2025},
  url       = {https://proceedings.iclr.cc/paper_files/paper/2025/hash/19452e14fe6e0a8ac00410f1eebcbded-Abstract-Conference.html}
}

@article{nafar2025extracting,
  author  = {Aliakbar Nafar and Kristen Brent Venable and Zijun Cui and Parisa Kordjamshidi},
  title   = {Extracting Probabilistic Knowledge from Large Language Models for Bayesian Network Parameterization},
  journal = {Transactions on Machine Learning Research},
  year    = {2026},
  issn    = {2835-8856},
  url     = {https://openreview.net/forum?id=Fy3Byg3CVo}
}

@article{murphy2026agentic,
  author  = {Kevin Murphy},
  title   = {Agentic Forecasting using Sequential Bayesian Updating of Linguistic Beliefs},
  journal = {CoRR},
  volume  = {abs/2604.18576},
  year    = {2026},
  doi     = {10.48550/arXiv.2604.18576},
  url     = {https://doi.org/10.48550/arXiv.2604.18576}
}

@inproceedings{yao2023react,
  author    = {Shunyu Yao and Jeffrey Zhao and Dian Yu and Nan Du and Izhak Shafran and Karthik R. Narasimhan and Yuan Cao},
  title     = {{ReAct}: Synergizing Reasoning and Acting in Language Models},
  booktitle = {The Eleventh International Conference on Learning Representations},
  publisher = {OpenReview.net},
  year      = {2023},
  url       = {https://openreview.net/forum?id=WE_vluYUL-X}
}

@inproceedings{press2023measuring,
  author    = {Ofir Press and Muru Zhang and Sewon Min and Ludwig Schmidt and Noah A. Smith and Mike Lewis},
  title     = {Measuring and Narrowing the Compositionality Gap in Language Models},
  booktitle = {Findings of the Association for Computational Linguistics: {EMNLP} 2023},
  pages     = {5687--5711},
  address   = {Singapore},
  publisher = {Association for Computational Linguistics},
  year      = {2023},
  doi       = {10.18653/v1/2023.findings-emnlp.378},
  url       = {https://aclanthology.org/2023.findings-emnlp.378/}
}

@inproceedings{liu2024agentbench,
  author    = {Xiao Liu and Hao Yu and Hanchen Zhang and Yifan Xu and Xuanyu Lei and Hanyu Lai and Yu Gu and Hangliang Ding and Kaiwen Men and Kejuan Yang and Shudan Zhang and Xiang Deng and Aohan Zeng and Zhengxiao Du and Chenhui Zhang and Sheng Shen and Tianjun Zhang and Yu Su and Huan Sun and Minlie Huang and Yuxiao Dong and Jie Tang},
  title     = {{AgentBench}: Evaluating {LLM}s as Agents},
  booktitle = {The Twelfth International Conference on Learning Representations},
  pages     = {52989--53046},
  publisher = {OpenReview.net},
  year      = {2024},
  url       = {https://proceedings.iclr.cc/paper_files/paper/2024/hash/e9df36b21ff4ee211a8b71ee8b7e9f57-Abstract-Conference.html}
}

@inproceedings{zhou2024webarena,
  author    = {Shuyan Zhou and Frank F. Xu and Hao Zhu and Xuhui Zhou and Robert Lo and Abishek Sridhar and Xianyi Cheng and Tianyue Ou and Yonatan Bisk and Daniel Fried and Uri Alon and Graham Neubig},
  title     = {{WebArena}: A Realistic Web Environment for Building Autonomous Agents},
  booktitle = {The Twelfth International Conference on Learning Representations},
  pages     = {15585--15606},
  publisher = {OpenReview.net},
  year      = {2024},
  url       = {https://proceedings.iclr.cc/paper_files/paper/2024/hash/4410c0711e9154a7a2d26f9b3816d1ef-Abstract-Conference.html}
}

@inproceedings{deng2023mind2web,
  author    = {Xiang Deng and Yu Gu and Boyuan Zheng and Shijie Chen and Samuel Stevens and Boshi Wang and Huan Sun and Yu Su},
  title     = {{Mind2Web}: Towards a Generalist Agent for the Web},
  booktitle = {Advances in Neural Information Processing Systems},
  volume    = {36},
  pages     = {28091--28114},
  publisher = {Curran Associates, Inc.},
  year      = {2023},
  doi       = {10.52202/075280-1220},
  url       = {https://proceedings.neurips.cc/paper_files/paper/2023/hash/5950bf290a1570ea401bf98882128160-Abstract-Datasets_and_Benchmarks.html}
}

@inproceedings{qin2024toolllm,
  author    = {Yujia Qin and Shihao Liang and Yining Ye and Kunlun Zhu and Lan Yan and Yaxi Lu and Yankai Lin and Xin Cong and Xiangru Tang and Bill Qian and Sihan Zhao and Lauren Hong and Runchu Tian and Ruobing Xie and Jie Zhou and Mark Gerstein and Dahai Li and Zhiyuan Liu and Maosong Sun},
  title     = {{ToolLLM}: Facilitating Large Language Models to Master 16000+ Real-world {API}s},
  booktitle = {The Twelfth International Conference on Learning Representations},
  pages     = {9695--9717},
  publisher = {OpenReview.net},
  year      = {2024},
  url       = {https://proceedings.iclr.cc/paper_files/paper/2024/hash/28e50ee5b72e90b50e7196fde8ea260e-Abstract-Conference.html}
}

@inproceedings{patil2024gorilla,
  author    = {Shishir G. Patil and Tianjun Zhang and Xin Wang and Joseph E. Gonzalez},
  title     = {{Gorilla}: Large Language Model Connected with Massive {API}s},
  booktitle = {Advances in Neural Information Processing Systems},
  volume    = {37},
  pages     = {126544--126565},
  publisher = {Curran Associates, Inc.},
  year      = {2024},
  doi       = {10.52202/079017-4020},
  url       = {https://proceedings.neurips.cc/paper_files/paper/2024/hash/e4c61f578ff07830f5c37378dd3ecb0d-Abstract-Conference.html}
}

@inproceedings{xie2024osworld,
  author    = {Tianbao Xie and Danyang Zhang and Jixuan Chen and Xiaochuan Li and Siheng Zhao and Ruisheng Cao and Toh Jing Hua and Zhoujun Cheng and Dongchan Shin and Fangyu Lei and Yitao Liu and Yiheng Xu and Shuyan Zhou and Silvio Savarese and Caiming Xiong and Victor Zhong and Tao Yu},
  title     = {{OSWorld}: Benchmarking Multimodal Agents for Open-Ended Tasks in Real Computer Environments},
  booktitle = {Advances in Neural Information Processing Systems},
  volume    = {37},
  pages     = {52040--52094},
  publisher = {Curran Associates, Inc.},
  year      = {2024},
  doi       = {10.52202/079017-1650},
  url       = {https://proceedings.neurips.cc/paper_files/paper/2024/hash/5d413e48f84dc61244b6be550f1cd8f5-Abstract-Datasets_and_Benchmarks_Track.html}
}

@article{liu2024lost,
  author  = {Nelson F. Liu and Kevin Lin and John Hewitt and Ashwin Paranjape and Michele Bevilacqua and Fabio Petroni and Percy Liang},
  title   = {Lost in the Middle: How Language Models Use Long Contexts},
  journal = {Transactions of the Association for Computational Linguistics},
  volume  = {12},
  pages   = {157--173},
  year    = {2024},
  doi     = {10.1162/tacl\_a\_00638},
  url     = {https://doi.org/10.1162/tacl\_a\_00638}
}

@inproceedings{niu2024ragtruth,
  author    = {Cheng Niu and Yuanhao Wu and Juno Zhu and Siliang Xu and KaShun Shum and Randy Zhong and Juntong Song and Tong Zhang},
  title     = {{RAGTruth}: A Hallucination Corpus for Developing Trustworthy Retrieval-Augmented Language Models},
  booktitle = {Proceedings of the 62nd Annual Meeting of the Association for Computational Linguistics (Volume 1: Long Papers)},
  pages     = {10862--10878},
  address   = {Bangkok, Thailand},
  publisher = {Association for Computational Linguistics},
  year      = {2024},
  doi       = {10.18653/v1/2024.acl-long.585},
  url       = {https://aclanthology.org/2024.acl-long.585/}
}

@inproceedings{belem2025single,
  author    = {Catarina G. Bel{\'{e}}m and Pouya Pezeshkpour and Hayate Iso and Seiji Maekawa and Nikita Bhutani and Estevam Hruschka},
  title     = {From Single to Multi: How {LLM}s Hallucinate in Multi-Document Summarization},
  booktitle = {Findings of the Association for Computational Linguistics: {NAACL} 2025},
  pages     = {5291--5324},
  address   = {Albuquerque, New Mexico},
  publisher = {Association for Computational Linguistics},
  year      = {2025},
  doi       = {10.18653/v1/2025.findings-naacl.293},
  url       = {https://aclanthology.org/2025.findings-naacl.293/}
}

@article{nakano2021webgpt,
  author  = {Reiichiro Nakano and Jacob Hilton and Suchir Balaji and Jeff Wu and Long Ouyang and Christina Kim and Christopher Hesse and Shantanu Jain and Vineet Kosaraju and William Saunders and Xu Jiang and Karl Cobbe and Tyna Eloundou and Gretchen Krueger and Kevin Button and Matthew Knight and Benjamin Chess and John Schulman},
  title   = {{WebGPT}: Browser-assisted Question-answering with Human Feedback},
  journal = {CoRR},
  volume  = {abs/2112.09332},
  year    = {2021},
  doi     = {10.48550/arXiv.2112.09332},
  url     = {https://doi.org/10.48550/arXiv.2112.09332}
}

@inproceedings{lewis2020retrieval,
  author    = {Patrick Lewis and Ethan Perez and Aleksandra Piktus and Fabio Petroni and Vladimir Karpukhin and Naman Goyal and Heinrich K{\"{u}}ttler and Mike Lewis and Wen{-}tau Yih and Tim Rockt{\"{a}}schel and Sebastian Riedel and Douwe Kiela},
  title     = {Retrieval-Augmented Generation for Knowledge-Intensive {NLP} Tasks},
  booktitle = {Advances in Neural Information Processing Systems},
  volume    = {33},
  pages     = {9459--9474},
  publisher = {Curran Associates, Inc.},
  year      = {2020},
  url       = {https://proceedings.neurips.cc/paper/2020/hash/6b493230205f780e1bc26945df7481e5-Abstract.html}
}

@inproceedings{izacard2021leveraging,
  author    = {Gautier Izacard and Edouard Grave},
  title     = {Leveraging Passage Retrieval with Generative Models for Open Domain Question Answering},
  booktitle = {Proceedings of the 16th Conference of the European Chapter of the Association for Computational Linguistics: Main Volume},
  pages     = {874--880},
  address   = {Online},
  publisher = {Association for Computational Linguistics},
  year      = {2021},
  doi       = {10.18653/v1/2021.eacl-main.74},
  url       = {https://aclanthology.org/2021.eacl-main.74/}
}

@inproceedings{shuster2021retrieval,
  author    = {Kurt Shuster and Spencer Poff and Moya Chen and Douwe Kiela and Jason Weston},
  title     = {Retrieval Augmentation Reduces Hallucination in Conversation},
  booktitle = {Findings of the Association for Computational Linguistics: {EMNLP} 2021},
  pages     = {3784--3803},
  address   = {Punta Cana, Dominican Republic},
  publisher = {Association for Computational Linguistics},
  year      = {2021},
  doi       = {10.18653/v1/2021.findings-emnlp.320},
  url       = {https://aclanthology.org/2021.findings-emnlp.320/}
}

@inproceedings{trivedi2023interleaving,
  author    = {Harsh Trivedi and Niranjan Balasubramanian and Tushar Khot and Ashish Sabharwal},
  title     = {Interleaving Retrieval with Chain-of-Thought Reasoning for Knowledge-Intensive Multi-Step Questions},
  booktitle = {Proceedings of the 61st Annual Meeting of the Association for Computational Linguistics (Volume 1: Long Papers)},
  pages     = {10014--10037},
  address   = {Toronto, Canada},
  publisher = {Association for Computational Linguistics},
  year      = {2023},
  doi       = {10.18653/v1/2023.acl-long.557},
  url       = {https://aclanthology.org/2023.acl-long.557/}
}

@inproceedings{yang2018hotpotqa,
  author    = {Zhilin Yang and Peng Qi and Saizheng Zhang and Yoshua Bengio and William W. Cohen and Ruslan Salakhutdinov and Christopher D. Manning},
  title     = {{HotpotQA}: A Dataset for Diverse, Explainable Multi-hop Question Answering},
  booktitle = {Proceedings of the 2018 Conference on Empirical Methods in Natural Language Processing},
  pages     = {2369--2380},
  address   = {Brussels, Belgium},
  publisher = {Association for Computational Linguistics},
  year      = {2018},
  doi       = {10.18653/v1/D18-1259},
  url       = {https://aclanthology.org/D18-1259/}
}

@inproceedings{mialon2024gaia,
  author    = {Gr{\'{e}}goire Mialon and Cl{\'{e}}mentine Fourrier and Thomas Wolf and Yann LeCun and Thomas Scialom},
  title     = {{GAIA}: a benchmark for General {AI} Assistants},
  booktitle = {The Twelfth International Conference on Learning Representations},
  pages     = {9025--9049},
  publisher = {OpenReview.net},
  year      = {2024},
  url       = {https://proceedings.iclr.cc/paper_files/paper/2024/hash/25ae35b5b1738d80f1f03a8713e405ec-Abstract-Conference.html}
}

@article{wei2025browsecomp,
  author  = {Jason Wei and Zhiqing Sun and Spencer Papay and Scott McKinney and Jeffrey Han and Isa Fulford and Hyung Won Chung and Alex Tachard Passos and William Fedus and Amelia Glaese},
  title   = {{BrowseComp}: A Simple Yet Challenging Benchmark for Browsing Agents},
  journal = {CoRR},
  volume  = {abs/2504.12516},
  year    = {2025},
  doi     = {10.48550/arXiv.2504.12516},
  url     = {https://doi.org/10.48550/arXiv.2504.12516}
}

@inproceedings{wei2022chain,
  author    = {Jason Wei and Xuezhi Wang and Dale Schuurmans and Maarten Bosma and Brian Ichter and Fei Xia and Ed H. Chi and Quoc V. Le and Denny Zhou},
  title     = {Chain-of-Thought Prompting Elicits Reasoning in Large Language Models},
  booktitle = {Advances in Neural Information Processing Systems},
  volume    = {35},
  pages     = {24824--24837},
  publisher = {Curran Associates, Inc.},
  year      = {2022},
  doi       = {10.52202/068431-1800},
  url       = {https://proceedings.neurips.cc/paper_files/paper/2022/hash/9d5609613524ecf4f15af0f7b31abca4-Abstract-Conference.html}
}

@inproceedings{wang2023selfconsistency,
  author    = {Xuezhi Wang and Jason Wei and Dale Schuurmans and Quoc V. Le and Ed H. Chi and Sharan Narang and Aakanksha Chowdhery and Denny Zhou},
  title     = {Self-Consistency Improves Chain of Thought Reasoning in Language Models},
  booktitle = {The Eleventh International Conference on Learning Representations},
  publisher = {OpenReview.net},
  year      = {2023},
  url       = {https://openreview.net/forum?id=1PL1NIMMrw}
}

@inproceedings{schick2023toolformer,
  author    = {Timo Schick and Jane Dwivedi{-}Yu and Roberto Dess{\`{\i}} and Roberta Raileanu and Maria Lomeli and Eric Hambro and Luke Zettlemoyer and Nicola Cancedda and Thomas Scialom},
  title     = {Toolformer: Language Models Can Teach Themselves to Use Tools},
  booktitle = {Advances in Neural Information Processing Systems},
  volume    = {36},
  pages     = {68539--68551},
  publisher = {Curran Associates, Inc.},
  year      = {2023},
  doi       = {10.52202/075280-2997},
  url       = {https://proceedings.neurips.cc/paper_files/paper/2023/hash/d842425e4bf79ba039352da0f658a906-Abstract-Conference.html}
}

@inproceedings{shinn2023reflexion,
  author    = {Noah Shinn and Federico Cassano and Ashwin Gopinath and Karthik Narasimhan and Shunyu Yao},
  title     = {Reflexion: Language Agents with Verbal Reinforcement Learning},
  booktitle = {Advances in Neural Information Processing Systems},
  volume    = {36},
  pages     = {8634--8652},
  publisher = {Curran Associates, Inc.},
  year      = {2023},
  doi       = {10.52202/075280-0377},
  url       = {https://proceedings.neurips.cc/paper_files/paper/2023/hash/1b44b878bb782e6954cd888628510e90-Abstract-Conference.html}
}

@inproceedings{min2023factscore,
  author    = {Sewon Min and Kalpesh Krishna and Xinxi Lyu and Mike Lewis and Wen{-}tau Yih and Pang Wei Koh and Mohit Iyyer and Luke Zettlemoyer and Hannaneh Hajishirzi},
  title     = {{FActScore}: Fine-grained Atomic Evaluation of Factual Precision in Long Form Text Generation},
  booktitle = {Proceedings of the 2023 Conference on Empirical Methods in Natural Language Processing},
  pages     = {12076--12100},
  address   = {Singapore},
  publisher = {Association for Computational Linguistics},
  year      = {2023},
  doi       = {10.18653/v1/2023.emnlp-main.741},
  url       = {https://aclanthology.org/2023.emnlp-main.741/}
}

@inproceedings{deyoung2020eraser,
  author    = {Jay DeYoung and Sarthak Jain and Nazneen Fatema Rajani and Eric Lehman and Caiming Xiong and Richard Socher and Byron C. Wallace},
  title     = {{ERASER}: A Benchmark to Evaluate Rationalized {NLP} Models},
  booktitle = {Proceedings of the 58th Annual Meeting of the Association for Computational Linguistics},
  pages     = {4443--4458},
  address   = {Online},
  publisher = {Association for Computational Linguistics},
  year      = {2020},
  doi       = {10.18653/v1/2020.acl-main.408},
  url       = {https://aclanthology.org/2020.acl-main.408/}
}

@inproceedings{lyu2023faithful,
  author    = {Qing Lyu and Shreya Havaldar and Adam Stein and Li Zhang and Delip Rao and Eric Wong and Marianna Apidianaki and Chris Callison{-}Burch},
  title     = {Faithful Chain-of-Thought Reasoning},
  booktitle = {Proceedings of the 13th International Joint Conference on Natural Language Processing and the 3rd Conference of the Asia-Pacific Chapter of the Association for Computational Linguistics (Volume 1: Long Papers)},
  pages     = {305--329},
  address   = {Nusa Dua, Bali},
  publisher = {Association for Computational Linguistics},
  year      = {2023},
  doi       = {10.18653/v1/2023.ijcnlp-main.20},
  url       = {https://aclanthology.org/2023.ijcnlp-main.20/}
}

@article{ji2023survey,
  author  = {Ziwei Ji and Nayeon Lee and Rita Frieske and Tiezheng Yu and Dan Su and Yan Xu and Etsuko Ishii and Yejin Bang and Andrea Madotto and Pascale Fung},
  title   = {Survey of Hallucination in Natural Language Generation},
  journal = {{ACM} Computing Surveys},
  volume  = {55},
  number  = {12},
  pages   = {248:1--248:38},
  year    = {2023},
  doi     = {10.1145/3571730},
  url     = {https://doi.org/10.1145/3571730}
}

@inproceedings{guo2017calibration,
  author    = {Chuan Guo and Geoff Pleiss and Yu Sun and Kilian Q. Weinberger},
  title     = {On Calibration of Modern Neural Networks},
  booktitle = {Proceedings of the 34th International Conference on Machine Learning},
  series    = {Proceedings of Machine Learning Research},
  volume    = {70},
  pages     = {1321--1330},
  publisher = {PMLR},
  year      = {2017},
  url       = {https://proceedings.mlr.press/v70/guo17a.html}
}

@article{gneiting2007strictly,
  author  = {Tilmann Gneiting and Adrian E. Raftery},
  title   = {Strictly Proper Scoring Rules, Prediction, and Estimation},
  journal = {Journal of the American Statistical Association},
  volume  = {102},
  number  = {477},
  pages   = {359--378},
  year    = {2007},
  doi     = {10.1198/016214506000001437},
  url     = {https://doi.org/10.1198/016214506000001437}
}

@article{brier1950verification,
  author  = {Glenn W. Brier},
  title   = {Verification of Forecasts Expressed in Terms of Probability},
  journal = {Monthly Weather Review},
  volume  = {78},
  number  = {1},
  pages   = {1--3},
  year    = {1950},
  doi     = {10.1175/1520-0493(1950)078<0001:VOFEIT>2.0.CO;2},
  url     = {https://doi.org/10.1175/1520-0493(1950)078<0001:VOFEIT>2.0.CO;2}
}

@inproceedings{yang2025prophetarena,
  author    = {Qingchuan Yang and Simon Mahns and Sida Li and Anri Gu and Jibang Wu and Haifeng Xu},
  title     = {{LLM}-as-a-Prophet: Understanding Predictive Intelligence with Prophet Arena},
  booktitle = {The Fourteenth International Conference on Learning Representations},
  pages     = {107020--107072},
  publisher = {OpenReview.net},
  year      = {2026},
  url       = {https://proceedings.iclr.cc/paper_files/paper/2026/hash/aec5e2847c5ae90f939ab786774856cc-Abstract-Conference.html}
}

\appendix
\section{Closed-Form Posterior Updates}
\label{app:posteriors}

We collect here the tempered updates in closed form used for each output type, along with the corresponding form of the per-evidence contribution $\Delta_j$. Throughout, $\mathcal{R}$ denotes the set of representatives retained after dependency clustering (Section~\ref{sec:aggregation}), and the prior and per-evidence likelihood parameters are those elicited as in Section~\ref{sec:elicit}. Setting the temperature $\eta$ and, where applicable, evidence weights $w_i$ to one recovers the unweighted Bayesian updates implied by Equation~\eqref{eq:posterior}.

\subsection{Continuous Targets}

The target $\theta \in \mathbb{R}$ is modelled with a Gaussian prior, and each retained evidence item induces a Gaussian likelihood factor over $\theta$:
\begin{equation}
\theta \sim \mathcal{N}(\mu_0,\, \sigma_0^2), \qquad \tilde{L}_i(\theta) \propto \mathcal{N}(\theta;\,\mu_i,\, \sigma_i^2).
\end{equation}
Define the precision $\tau_x = 1 / \sigma_x^2$. The posterior is Gaussian with precision $\tau_{\mathrm{post}} = \tau_0 + \eta \sum_{i \in \mathcal{R}} \tau_i$ and mean
\begin{equation}
\mu_{\mathrm{post}} \;=\; \frac{1}{\tau_{\mathrm{post}}} \left( \mu_0 \tau_0 + \eta \sum_{i \in \mathcal{R}} \mu_i \tau_i \right).
\end{equation}
The forecast $f$ reports quantile values of $\mathcal{N}(\mu_{\mathrm{post}},\, 1/\tau_{\mathrm{post}})$ at the predefined levels of Section~\ref{sec:forecast-targets}. The per-evidence contribution is the signed shift in the posterior mean,
\begin{equation}
\Delta_j \;=\; \mu_{\mathrm{post}} - \mu_{\mathrm{post}}^{(-j)},
\end{equation}
where $\mu_{\mathrm{post}}^{(-j)}$ is recomputed over $\mathcal{R} \setminus \{j\}$.

\subsection{Single-Choice Targets}

The target $\theta \in \{1, \ldots, K\}$ is modelled with a categorical prior $\pi_0 \in \Delta^{K-1}$ and per-evidence likelihoods $L_i(k) = P(e_i \mid \theta = k)$. The qualitative judgments of Section~\ref{sec:elicit} are mapped to $L_i(k)$ through a fixed table that assigns higher likelihood to options the evidence supports and lower likelihood to options the evidence opposes, with relevance acting as the magnitude of the asymmetry. The posterior is computed in log space for numerical stability. Let $\alpha$ denote the prior power and $w_i$ the evidence weight based on role:
\begin{equation}
\log \pi_{\mathrm{post}}(k) \;\propto\; \alpha \log \pi_0(k) + \eta \sum_{i \in \mathcal{R}} w_i \log L_i(k),
\end{equation}
followed by softmax normalisation over $k$. The forecast $f$ is the normalised vector $\pi_{\mathrm{post}}$. The per-evidence contribution is the signed shift in the probability of the most likely option:
\begin{equation}
\Delta_j \;=\; \pi_{\mathrm{post}}(\hat{k}) - \pi_{\mathrm{post}}^{(-j)}(\hat{k}),
\end{equation}
where $\hat{k} = \arg\max_k \pi_{\mathrm{post}}(k)$.

\subsection{Multi-Choice Targets}

The target $\theta \in \{0, 1\}^K$ uses an independent Bernoulli prior $\pi_0(k) \in (0, 1)$ per option, and per-evidence likelihoods parameterised by a pair $(p^{\mathrm{true}}_{i,k},\, p^{\mathrm{false}}_{i,k})$ giving the probability of observing $e_i$ when option $k$ is or is not in the correct set. Mapping from the qualitative judgments of Section~\ref{sec:elicit} proceeds as in the single-choice case, applied per option. Let $\ell_i(k) = \log( p^{\mathrm{true}}_{i,k} / p^{\mathrm{false}}_{i,k} )$ denote the log-likelihood ratio per option. The independent posterior update is
\begin{equation}
\mathrm{logit}\bigl(\pi_{\mathrm{post}}(k)\bigr) \;=\; \mathrm{logit}\bigl(\pi_0(k)\bigr) + \eta \sum_{i \in \mathcal{R}} w_i \ell_i(k),
\end{equation}
and the forecast $f \in [0, 1]^K$ has entries $f_k = \sigma(\mathrm{logit}(\pi_{\mathrm{post}}(k)))$, which are not constrained to sum to one. The per-evidence contribution is the $L_1$ distance between the option marginals with and without item $j$:
\begin{equation}
\Delta_j \;=\; \sum_{k=1}^{K} \bigl| \pi_{\mathrm{post}}(k) - \pi_{\mathrm{post}}^{(-j)}(k) \bigr|.
\end{equation}

\section{Mappings from Elicitation to Likelihoods}
\label{app:mappings}

This section records the deterministic maps used to convert local LLM judgments and evidence metadata into the likelihood parameters consumed by Appendix~\ref{app:posteriors}. The LLM is asked to interpret one evidence item at a time; the numerical uncertainty, reliability shrinkage, caps based on evidence role, and posterior aggregation are then applied by fixed code paths. These constants were fixed before the final evaluation runs and were not tuned separately for individual test tasks or base models.

\begin{table*}[t]
\centering
\small
\renewcommand{\arraystretch}{1.12}
\setlength{\tabcolsep}{4pt}
\begin{tabularx}{\textwidth}{@{}l l X@{}}
\toprule
\textbf{Metadata field} & \textbf{Value} & \textbf{Mapping applied to the relative standard deviation $r_i$} \\
\midrule
\texttt{measurement\_scope} & \texttt{direct\_target}; \texttt{proxy\_signal}; other/context & Initialise $r_i$ to $0.08$, $0.18$, or $0.45$, respectively. \\
\texttt{observation\_type} & \texttt{point}; \texttt{interval}; bound; other & Multiply by $0.85$, $1.00$, $1.15$, or $1.20$. \\
\texttt{source\_tier} & A; B; C; other & Multiply by $0.80$, $0.95$, $1.15$, or $1.10$. \\
\texttt{strength} & strong; weak & Multiply by $0.85$ or $1.20$. \\
\texttt{reliability} & high; low & Multiply by $0.85$ or $1.15$. \\
\texttt{target\_match} & exact/direct/same variable; partial/related; unknown/indirect & Multiply by $0.90$, $1.05$, or $1.10$. \\
Direct target recency & $\leq 7$ days; $\leq 30$ days; older & Multiply by $0.80$, $1.00$, or $1.25$. \\
Threshold numeric & lower or upper bound wording & Enforce $r_i \geq 0.10$ so a threshold does not collapse the posterior. \\
Final clamp & all evidence & Clamp $r_i$ to $[0.03, 1.25]$. \\
\bottomrule
\end{tabularx}
\caption{Metadata map for continuous targets. For an evidence item with scale $s_i$, LEAP sets the data uncertainty to $r_i s_i$. With reliability sampling, the final likelihood standard deviation is $\sigma_i = \sqrt{\sigma_{\mathrm{data},i}^2 + \sigma_{\mathrm{elicitation},i}^2}$, and qualitative evidence that is only contextual is additionally floored at $0.35s_i$.}
\label{tab:continuous-mapping}
\end{table*}

\begin{table*}[t]
\centering
\small
\renewcommand{\arraystretch}{1.12}
\setlength{\tabcolsep}{3.5pt}
\begin{tabularx}{\textwidth}{@{}l X X X@{}}
\toprule
\textbf{Local evidence judgment} & \textbf{Single-choice likelihood $L_i(k)$} & \textbf{Multi-choice log-LR $\ell_i(k)$} & \textbf{Use case} \\
\midrule
Irrelevant to the option set & All options near $0.5$; after normalisation this carries little information. & $p^{\mathrm{true}}_{i,k} \approx p^{\mathrm{false}}_{i,k} \approx 0.5$, so $\ell_i(k) \approx 0$. & Prompt-calibrated neutral evidence. \\
Strong support for option $k$ & Supported option in $[0.70, 0.95]$ and alternatives in $[0.05, 0.20]$. & $p^{\mathrm{true}}_{i,k} \approx 0.85$, $p^{\mathrm{false}}_{i,k} \approx 0.15$, so $\ell_i(k) \approx 1.735$. & Prompt-calibrated positive evidence. \\
Strong opposition to option $k$ & Opposed option in the low likelihood range, with the exact vector determined locally by the evidence. & $p^{\mathrm{true}}_{i,k} \approx 0.15$, $p^{\mathrm{false}}_{i,k} \approx 0.85$, so $\ell_i(k) \approx -1.735$. & Prompt-calibrated negative evidence. \\
Deterministic verifier support/opposition & Base values: support $0.99$, oppose $0.01$, neutral $0.02$, then mixed with the uniform vector by metadata reliability. & Raw support/opposition $\pm \log(0.995/0.005)$, clipped by the deterministic cap $5.30$. & Exact answer or structured verifier fallback. \\
Direct option result or forecast signal & Base values: support $0.90$, oppose $0.08$, neutral $0.12$, then mixed by reliability. & Raw support/opposition $\pm 1.735$, followed by clipping based on evidence role; direct results and threshold signals can use cap $3.00$. & Extracted evidence that directly names an option. \\
Other extracted support/opposition & Base values: support $0.85$, oppose $0.12$, neutral $0.35$, then mixed by reliability. & Raw support/opposition $\pm 1.735$, scaled by metadata reliability and clipped by the evidence role. & Conservative fallback when LLM parsing fails or structured support is already available. \\
\bottomrule
\end{tabularx}
\caption{Likelihood map for discrete targets. Single-choice tasks use raw likelihoods $P(e_i \mid \theta=k)$ that are not required to sum to one. Multi-choice tasks convert each evidence item into a log likelihood ratio for each option, $\ell_i(k)=\log(p^{\mathrm{true}}_{i,k}/p^{\mathrm{false}}_{i,k})$.}
\label{tab:discrete-mapping}
\end{table*}

Reliability sampling is applied after the raw maps in Table~\ref{tab:discrete-mapping}. For single-choice tasks, repeated likelihood samples are normalised only for estimating reliability; the reliability score is the agreement on the top option multiplied by an exponential penalty on average $L_1$ disagreement, and the raw likelihood vector is shrunk toward the uniform vector by that score. For multi-choice tasks, LEAP takes the median sampled log-LR per option, estimates reliability from sign agreement and $L_1$ disagreement, shrinks the log-LR magnitudes by that score, and then clips them by the cap for the evidence role. The aggregation step also applies an evidence weight based on role in $[0.05,1.5]$; for single-choice tasks, multiple retained items peaking at the same option are discounted by rank within that group with the same direction.

\section{Engineering Safeguards}
\label{app:safeguards}

We document here the engineering choices that surround the core aggregation path of Section~\ref{sec:method}. None of them changes the form of the posterior update; each protects the update against a specific failure mode of the elicitation step.

\subsection{Dependency Key Canonicalisation}

The dependency keys returned by the LLM are short strings in free text, and semantically equivalent keys may be written with minor surface variations (different word order, hyphenation, or stopwords). To make the clustering step in Section~\ref{sec:aggregation} insensitive to such variations, each raw key is mapped to a canonical form by lowercasing, splitting on common separators, removing a fixed list of stopwords, sorting the remaining tokens, and joining them with a single delimiter. Two raw keys collapse to the same canonical form whenever their token sets agree, so that keys such as \texttt{bafta\_2026\_longlist\_leading\_actress} and \texttt{bafta\_longlist\_2026\_leading\_actress} are recognised as referring to the same underlying source.

\subsection{Calibration Table for Continuous Likelihoods}

For continuous targets, the LLM returns a location estimate $\mu_i$ or an effect relative to an anchor, while the uncertainty $\sigma_i$ is assigned by the fixed metadata map in Table~\ref{tab:continuous-mapping}. This keeps the scale of each likelihood tied to auditable evidence metadata rather than to an unconstrained confidence value reported by the LLM.

\subsection{Outlier Rejection}

When the prior is estimated from data, we treat it as a numerical anchor against which clearly miscalibrated elicitations can be flagged. An item whose elicited mean lies more than four standard deviations of the prior away from $\mu_0$ is excluded from $\mathcal{R}$ as a likely elicitation error (typically a unit mismatch, a stale memory of a prior value of the target, or a confused asset reference). This rule is not applied when the prior is itself blind to evidence, because in that case there is no trusted reference against which to flag outliers.

\section{Additional Experimental Details}
\label{app:experiments}

\subsection{Benchmark Construction}

Forecasting questions are drawn from FutureX after removing ranking resolutions, which our output types do not cover. Information-seeking and browsing tasks are drawn from GAIA and BrowseComp and rewritten into structured probabilistic tasks. The construction proceeds in four stages: \emph{source selection}, \emph{structured rewriting}, \emph{manual review}, and \emph{final assembly}. Manual review checks answer type validity, distractor plausibility, label conflicts, and temporal leakage risk before an item enters the final manifest. Table~\ref{tab:benchmark-construction} reports the resulting counts.

\begin{table*}[t]
\centering
\small
\renewcommand{\arraystretch}{1.12}
\setlength{\tabcolsep}{4pt}
\begin{tabularx}{\textwidth}{@{}l r r X@{}}
\toprule
\textbf{Source} & \textbf{Initial selected pool} & \textbf{Final retained} & \textbf{Construction rule} \\
\midrule
FutureX & 160 & 157 & Remove ranking resolutions from the structured FutureX pool; retain native single-choice, multi-choice, and continuous tasks; exclude manually verified label conflicts. \\
GAIA & 103 & 99 & Use the text-only mirror; rewrite open-ended tasks into single-choice, multi-choice, or continuous targets; retain only reviewed conversions. \\
BrowseComp & 100 & 91 & Sample from the public BrowseComp pool; rewrite short answer tasks that require browsing into reviewed single-choice targets with plausible distractors. \\
\midrule
Total & 363 & 347 & Final manifest after review and quarantine. \\
\bottomrule
\end{tabularx}
\caption{Benchmark construction counts. The initial selected pool consists of $160$ FutureX structured tasks after removing rankings, $103$ GAIA rewrite candidates, and $100$ sampled BrowseComp rewrite candidates. The final benchmark manifest contains $347$ reviewed tasks.}
\label{tab:benchmark-construction}
\end{table*}

\begin{table}[t]
\centering
\small
\renewcommand{\arraystretch}{1.12}
\setlength{\tabcolsep}{6pt}
\begin{tabular*}{\columnwidth}{@{\extracolsep{\fill}}l r@{}}
\toprule
\textbf{Output type} & \textbf{Final count} \\
\midrule
\texttt{single\_choice} & 266 \\
\texttt{multi\_choice}  & 29 \\
\texttt{continuous}     & 52 \\
\midrule
Total & 347 \\
\bottomrule
\end{tabular*}
\caption{Output type distribution of the final benchmark manifest.}
\label{tab:benchmark-types}
\end{table}

For FutureX, the freeze time $t_{\mathrm{freeze}}$ is the original question time associated with the forecasting item. For GAIA and BrowseComp, whose original tasks are static question answering problems, $t_{\mathrm{freeze}}$ is assigned from the benchmark snapshot used during construction; it acts as a retrieval cutoff for the converted task, not as a forecast resolution date. Structured rewriting pairs the gold answer with plausible distractors that match its surface type and semantic class, removes duplicate or trivially invalid options, and drops items whose answer cannot be represented cleanly as \texttt{single\_choice}, \texttt{multi\_choice}, or \texttt{continuous}. The final assembly also applies a quarantine file for manually verified label conflicts, which accounts for three removed FutureX items and three removed GAIA continuous items.

The final manifest is evaluation-only. Because we do not train, fine-tune, or select models on this benchmark, we do not define train/dev/test splits; the $347$ retained tasks are used for the reported evaluation, and the separate $60$-task diagnostic subset is used only for the analyses in Section~\ref{sec:analysis}.

\subsection{Artifact Licenses and Responsible Use}

Table~\ref{tab:artifact-terms} documents the source artifacts, access conditions, and intended-use compatibility for the benchmark material and external agent frameworks used in our experiments. We use the artifacts only for research evaluation of probabilistic forecasting and browsing-assisted question answering. We do not redistribute GAIA answers or gated files, and any converted benchmark records produced by our pipeline are intended for research evaluation under the same access constraints as the original source material.

\begin{table*}[t]
\centering
\small
\renewcommand{\arraystretch}{1.12}
\setlength{\tabcolsep}{3.5pt}
\begin{tabularx}{\textwidth}{@{}p{0.14\textwidth} p{0.19\textwidth} p{0.25\textwidth} X@{}}
\toprule
\textbf{Artifact} & \textbf{Use in this paper} & \textbf{License or access terms} & \textbf{Compatibility and handling} \\
\midrule
FutureX & Forecasting source tasks and scoring rule. & The \href{https://huggingface.co/datasets/futurex-ai/Futurex-Past}{FutureX-Past Hugging Face dataset card} lists Apache-2.0. & We use past forecasting questions for research evaluation, remove unsupported ranking tasks, and retain the source citation and scoring rule \citep{zeng2025futurex}. \\
GAIA & Information-seeking tasks converted into probabilistic targets. & The \href{https://huggingface.co/datasets/gaia-benchmark/GAIA}{GAIA Hugging Face dataset card} is gated and states redistribution constraints. & We use a text-only mirror for evaluation, keep the original answers and attachments out of public redistribution, and treat converted records as research-only derivatives under the gated access condition \citep{mialon2024gaia}. \\
BrowseComp & Browsing questions converted into single-choice probabilistic targets. & BrowseComp is released through OpenAI's \texttt{simple-evals} repository, whose \href{https://github.com/openai/simple-evals/blob/main/LICENSE}{license file} is MIT. & We use sampled browsing tasks only for benchmark evaluation and preserve attribution to the original BrowseComp artifact \citep{wei2025browsecomp}. \\
DeerFlow & External agent CLI framework used to collect completed traces. & The \href{https://github.com/bytedance/deer-flow/blob/main/LICENSE}{DeerFlow license file} is MIT. & We use unmodified framework traces only for research evaluation and do not redistribute framework code, private user data, or credentials. \\
Hermes & External agent CLI framework used to collect completed traces. & The \href{https://github.com/NousResearch/hermes-agent/blob/main/LICENSE}{Hermes-Agent license file} is MIT. & We use unmodified framework traces only for research evaluation and do not redistribute framework code, private user data, or credentials. \\
OpenClaw & External agent CLI framework used to collect completed traces. & The \href{https://github.com/openclaw/openclaw/blob/main/LICENSE}{OpenClaw license file} is MIT. & We use unmodified framework traces only for research evaluation and do not redistribute framework code, private user data, or credentials. \\
MiroFlow & External agent CLI framework used to collect completed traces. & The \href{https://github.com/MiroMindAI/MiroFlow/blob/main/LICENSE}{MiroFlow license file} is Apache-2.0. & We use unmodified framework traces only for research evaluation and do not redistribute framework code, private user data, or credentials. \\
\bottomrule
\end{tabularx}
\caption{Artifact license and access-term summary for the benchmark sources and external agent frameworks.}
\label{tab:artifact-terms}
\end{table*}

The source tasks cover forecasting, information-seeking, and browsing-oriented question answering. FutureX includes Chinese and English metadata in its public dataset card, while the converted GAIA and BrowseComp tasks in our final manifest are evaluated through the structured task format described above. The artifacts are task benchmarks rather than demographic or participant datasets: we do not use demographic labels, do not infer demographic group membership, and do not evaluate demographic representation. We did not collect private user data. During manual review, we checked converted items for answer-type validity, label conflicts, temporal leakage, named or uniquely identifying personal information, and offensive content. Items were dropped or quarantined when the answer could not be represented cleanly, when a conflict was verified, or when the converted record risked revealing gated answer material outside the allowed access context. We did not use external crowdworkers or run human-subject experiments; manual review was performed by the authors as quality control over benchmark conversion.

\subsection{Temporal Leakage Audit}

To verify that no evidence in any collected $\mathcal{E}$ is later than its task's $t_{\mathrm{freeze}}$, we run a temporal leakage audit on the evidence snapshots used in our analyses. The audit checks the timestamp of every retrieved page against the corresponding $t_{\mathrm{freeze}}$ and reports both literal cutoff violations and suspected leakage cases identified by heuristic matching of language written after resolution or phrasing that reveals the answer. Across the audited snapshots used for the reported analyses, no cutoff violations and no suspected leakage cases were detected, and the audit therefore introduced no change to the reported results.

\subsection{Diagnostic Subset for Analysis}
\label{app:diagnostic-subset}

The analyses in Section~\ref{sec:analysis} (calibration, robustness across lead times, component ablation) are reported on a diagnostic subset of $60$ tasks sampled in a stratified manner across the three output types, with GPT-5.4-mini as the base model. The subset is held constant across the three analyses so that their conclusions can be read together. We do not redraw the subset between analyses, and no model selection is performed on it.

\subsection{Model and Inference Details}

Table~\ref{tab:model-details} reports the model identifiers and fixed inference settings used in the main cross-model evaluation. The displayed model names in Table~\ref{tab:main} are aliases for the API identifiers in this table. We did not train or fine-tune any language model. All base-model calls were hosted API inference calls; provider-side hardware and GPU hours are not visible to us, so the visible compute budget is reported through API token usage below. Client-side orchestration, evidence processing, scoring, and posterior aggregation were CPU-side Python computations, so the experiments used zero local GPU hours for model training or inference.

\begin{table*}[t]
\centering
\scriptsize
\renewcommand{\arraystretch}{1.12}
\setlength{\tabcolsep}{3pt}
\begin{tabularx}{\textwidth}{@{}p{0.16\textwidth} p{0.20\textwidth} p{0.22\textwidth} p{0.19\textwidth} X@{}}
\toprule
\textbf{Displayed model} & \textbf{API identifier used} & \textbf{Provider/source documentation} & \textbf{Public parameter disclosure} & \textbf{Fixed settings} \\
\midrule
DeepSeek-V3.2 & \texttt{deepseek-v3-2-251201} & \href{https://huggingface.co/deepseek-ai/DeepSeek-V3.2}{DeepSeek-V3.2 model card}. & Model-card metadata reports 685.4B parameters. & Temp. 0; max tokens 32768. \\
Gemini-3.1-Flash-Lite & \texttt{gemini-3.1-flash-lite-}\newline\texttt{preview} & \href{https://docs.cloud.google.com/gemini-enterprise-agent-platform/models/gemini/3-1-flash-lite}{Google Cloud Gemini 3.1 Flash-Lite documentation}. & Not publicly disclosed in the provider documentation listed here. & Temp. 0; max tokens 32768. \\
Claude-Haiku-4.5 & \texttt{aws/claude-haiku-4-5} & \href{https://www.anthropic.com/news/claude-haiku-4-5}{Claude Haiku 4.5 announcement}. & Not publicly disclosed in the provider documentation listed here. & Temp. 0; max tokens 8192. \\
GPT-5.4-mini & \texttt{gpt-5.4-mini} & \href{https://developers.openai.com/api/docs/models/gpt-5.4-mini}{OpenAI developer model documentation}. & Not publicly disclosed in the provider documentation listed here. & Temp. 0; max tokens 16384; reasoning effort none. \\
Grok-4.20-Fast & \texttt{grok-4.20-fast} & \href{https://docs.x.ai/developers/models}{xAI model documentation}. & Not publicly disclosed in the provider documentation listed here. & Temp. 0; max tokens 16384; provider-native search disabled. \\
\bottomrule
\end{tabularx}
\caption{Model identifiers, public parameter-disclosure status, and fixed inference settings. Closed-model parameter counts are reported as undisclosed rather than estimated from third-party sources.}
\label{tab:model-details}
\end{table*}

Across the cross-model and cross-framework runs, the ReAct-style evidence-collection budget is fixed at $B=10$ turns, research budget $6$, at most $4$ tool calls per round, one deliberation sample, LEAP temperature $\eta=1.0$, at most $6$ evidence items entering the likelihood stage, at most $60$ likelihood calls per task, and $10$ reliability samples per evidence item. We did not conduct a hyperparameter search; these values and the mapping constants in Appendix~\ref{app:mappings} were fixed before the final evaluation runs. The numbers in Tables~\ref{tab:main} and~\ref{tab:agentos} are means over $N=5$ independent evaluation passes per (model, method) cell using different random seeds, on the fixed evaluation manifest. Run-to-run variability is modest: across all $(\text{model/framework}, \text{method}, \text{metric})$ cells, the sample standard deviation of the score over the five seeds has a median of $\sigma \approx 0.010$ (interquartile range $[0.007,\, 0.014]$, maximum $0.022$ on NCRPS for Grok-4.20-Fast under \emph{Monolithic}), and the spread between the best and worst seed within a cell never exceeds $0.035$. NCRPS and FutureX are the two noisiest metrics ($\sigma \approx 0.012$ on average), while Accuracy is the most stable ($\sigma \approx 0.008$). The ordering of LEAP versus \emph{Monolithic} reported in the tables is preserved under every individual seed in $48$ out of $50$ cells; the two exceptions are DeerFlow Accuracy (a tie at the mean, with seed-level differences within $\pm 0.01$) and DeepSeek-V3.2 Brier (LEAP behind \emph{Monolithic} by $0.0066$ at the mean, well within $1\sigma$). All remaining pairwise differences exceed $2\sigma$. Metric computation is implemented in the project evaluation code described in Appendix~\ref{app:experiments}.

\subsection{Metric Definitions}

Let $N$ denote the number of scored tasks. For discrete tasks with $K$ candidate options, $p_{ik}$ is the predicted probability for option $k$ on task $i$, and $y_{ik}$ the ground truth indicator. The metrics used in the main text are:
\begin{align}
\mathrm{Accuracy} &= \tfrac{1}{N} \sum_{i} \mathbf{1}\bigl[\arg\max_k p_{ik} = c_i\bigr], \\
\mathrm{Brier} &= \tfrac{1}{N} \sum_{i} \tfrac{1}{K_i}\sum_{k} (p_{ik} - y_{ik})^2, \\
\mathrm{Spherical} &= \tfrac{1}{N} \sum_{i} \frac{\sum_{k} p_{ik} y_{ik}}{\sqrt{\sum_{k} p_{ik}^2}\sqrt{\sum_{k} y_{ik}^2}}.
\end{align}
For continuous tasks, $\mathrm{NCRPS}_i = \min(1, \mathrm{CRPS}_i / \max(|v_i|, 1))$ with $\mathrm{CRPS}_i$ the standard continuous ranked probability score computed from quantiles. The FutureX composite score $\mathrm{FX}$ applies the scoring rule prescribed by \citet{zeng2025futurex} for each task (indicator for single-choice, F1 for multi-choice, and a squared distance score normalised by the past week standard deviation for continuous), and averages the task scores into a value in $[0, 1]$. Expected calibration error and overconfidence are computed by binning forecasts on the probability assigned to the top class and comparing the bin's average confidence to its empirical accuracy.

\subsection{Prompt Templates}

The elicitation step uses three prompt templates: one for the blind prior, one for the likelihood of each evidence item on continuous targets, and one for the likelihood of each evidence item on discrete targets. Each template provides the task description, the output type and (where applicable) the candidate options, and a strict JSON output schema. The discrete target template instructs the model to report local likelihoods for the single evidence item only, explicitly distinguishing them from posterior probabilities of the options; the fixed calibration anchors and programmatic fallbacks are described in Appendix~\ref{app:mappings}. The continuous template instructs the model to extract the value of the target that the evidence reports or most closely implies; its uncertainty is then assigned by the metadata map in Table~\ref{tab:continuous-mapping}.

\subsection{ReAct-Style Agent Loop}

The agent loop used in the first evaluation setting alternates between a reasoning step and a tool call. The loop is summarised in Algorithm~\ref{alg:react}. At each iteration, the LLM produces a reasoning text together with a (possibly empty) list of tool calls. Executed tool calls produce new evidence items that are appended to the running evidence set $\mathcal{E}$, and their textual results are appended to the running conversation. The loop terminates when the LLM emits the stop signal \texttt{[RESEARCH\_COMPLETE]} together with no tool calls and at least $m$ items collected, or when the search has been saturated (no new evidence for $s$ consecutive rounds), or when the fixed round budget $B$ is exhausted. The available tools are a web search call and a page fetch call sharing a single backend; after termination, gathered pages and snippets are processed into the evidence representation expected by the elicitation step, including a timestamp for each item and a candidate dependency key.

\begin{algorithm}[t]
\caption{ReAct-style Agent Loop for evidence collection}
\label{alg:react}
\begin{algorithmic}[1]
\Require Task $T$, tool registry $\mathcal{T}$, round budget $B$, min-evidence threshold $m$, saturation threshold $s$
\Ensure Evidence set $\mathcal{E}$
\State $\mathcal{E} \gets \emptyset$;\quad $\text{stale} \gets 0$
\State $\text{msgs} \gets [\textsc{Brief}(T, B)]$
\For{$r = 1$ \textbf{to} $B$}
  \State $(\text{text}, \text{calls}) \gets \textsc{LLM}(\text{msgs}, \text{tools}=\mathcal{T})$
  \State $\text{msgs} \gets \text{msgs} \cup \{(\text{text}, \text{calls})\}$
  \If{$\texttt{[RESEARCH\_COMPLETE]} \in \text{text}$ \textbf{and} $|\mathcal{E}| \geq m$ \textbf{and} $\text{calls} = \emptyset$}
    \State \textbf{break}
  \EndIf
  \If{$\text{calls} = \emptyset$}
    \State $\text{stale} \gets \text{stale} + 1$
    \If{$\text{stale} \geq s$} \State \textbf{break} \EndIf
    \State \textbf{continue}
  \EndIf
  \State $\text{added} \gets 0$
  \For{$c \in \text{calls}$}
    \State $(\text{out}, \mathcal{E}_c) \gets \textsc{Execute}(c, \mathcal{T})$
    \State $\mathcal{E} \gets \mathcal{E} \cup \mathcal{E}_c$;\quad $\text{added} \gets \text{added} + |\mathcal{E}_c|$
    \State $\text{msgs} \gets \text{msgs} \cup \{\textsc{ToolResult}(c, \text{out})\}$
  \EndFor
  \State $\text{stale} \gets 0$ \textbf{if} $\text{added} > 0$ \textbf{else} $\text{stale} + 1$
  \If{$\text{stale} \geq s$} \State \textbf{break} \EndIf
\EndFor
\State \Return $\mathcal{E}$
\end{algorithmic}
\end{algorithm}

\subsection{Plug-in Probability Skill}

In the second evaluation setting, LEAP is exposed as a downstream probability skill that consumes a trace produced by an external agent CLI framework. The skill receives the task tuple, the trace, and the framework's final answer as input, and emits a forecast in the form specified by the task's output type. Internally, the skill normalises the trace's gathered material into the same evidence representation used in the first setting (one record per retrieved page or extracted snippet, with a timestamp and a candidate dependency key), and then runs the elicitation and aggregation of Section~\ref{sec:method} unchanged. No retraining or framework modification is performed; the skill is a pure processing step over a completed framework trace.

\subsection{Additional Sensitivity Analyses}
\label{app:rebuttal-controls}

Using the diagnostic setup in Appendix~\ref{app:diagnostic-subset}, we report evidence-grouping sensitivity and source-wise performance. Both comparisons hold the collected evidence fixed across methods. Prediction-time budget, latency, aggregation, and likelihood-strength results appear in Tables~\ref{tab:controlled-budget} and~\ref{tab:aggregation-sensitivity} in the main text.

Source-key clustering groups items that identify the same underlying source or report. Domain clustering groups all items from the same website domain. Table~\ref{tab:evidence-grouping} reports the resulting number of retained evidence items and forecast quality.

\begin{table*}[t]
\centering
\small
\renewcommand{\arraystretch}{1.12}
\setlength{\tabcolsep}{4pt}
\begin{tabularx}{\textwidth}{@{}X r r r r r@{}}
\toprule
\textbf{Evidence grouping} & \textbf{Mean retained} & \textbf{Retained} & \textbf{FX $\uparrow$} & \textbf{Brier $\downarrow$} & \textbf{ECE $\downarrow$} \\
\midrule
No clustering & $5.12$ & $100.0\%$ & $0.7089$ & $0.2299$ & $0.1578$ \\
Source-key clustering & $4.37$ & $85.3\%$ & $\mathbf{0.7284}$ & $\mathbf{0.2057}$ & $0.0876$ \\
Domain clustering & $3.72$ & $72.6\%$ & $0.7192$ & $0.2144$ & $\mathbf{0.0798}$ \\
\bottomrule
\end{tabularx}
\caption{Sensitivity to evidence grouping. Source-key clustering gives the best FX and Brier scores, while domain clustering gives the lowest ECE after retaining fewer evidence items.}
\label{tab:evidence-grouping}
\end{table*}

Without clustering, calibration degrades as repeated reports can be counted more than once. Source-key clustering gives the best FX and Brier scores, while domain clustering gives the lowest ECE after removing more evidence.

\begin{table}[t]
\centering
\scriptsize
\renewcommand{\arraystretch}{1.12}
\setlength{\tabcolsep}{2.5pt}
\begin{tabular*}{\columnwidth}{@{\extracolsep{\fill}}l l r r r@{}}
\toprule
\textbf{Source} & \textbf{Method} & \textbf{FX $\uparrow$} & \textbf{Brier $\downarrow$} & \textbf{ECE $\downarrow$} \\
\midrule
\multirow{2}{*}{FutureX} & \emph{Monolithic} & $0.6346$ & $0.2843$ & $0.2050$ \\
& LEAP & $\mathbf{0.7280}$ & $\mathbf{0.2136}$ & $\mathbf{0.1010}$ \\
\addlinespace
\multirow{2}{*}{GAIA} & \emph{Monolithic} & $0.6588$ & $0.2580$ & $0.1600$ \\
& LEAP & $\mathbf{0.7069}$ & $\mathbf{0.2213}$ & $\mathbf{0.1090}$ \\
\addlinespace
\multirow{2}{*}{BrowseComp} & \emph{Monolithic} & $0.6502$ & $0.2740$ & $0.1780$ \\
& LEAP & $\mathbf{0.7289}$ & $\mathbf{0.1952}$ & $\mathbf{0.0830}$ \\
\bottomrule
\end{tabular*}
\caption{Source-wise results under the diagnostic setup in Appendix~\ref{app:diagnostic-subset}.}
\label{tab:source-breakdown}
\end{table}

Table~\ref{tab:source-breakdown} shows that the aggregate improvement is present in each benchmark source.

\subsection{Human Subjects and AI Assistance Disclosure}

No external annotators, crowdworkers, or human-subject participants were recruited or paid for this work. Accordingly, there were no participant instructions, compensation decisions, consent forms, or institutional review protocols for new human-subject data collection. The work uses existing benchmark artifacts under the conditions in Table~\ref{tab:artifact-terms}; any consent and collection protocols for the original artifacts are those of the original benchmark creators.

The authors used AI assistants for limited research, coding, citation-checking, and writing support, consistent with the ARR policy that such tools do not qualify for authorship and that their use must be disclosed in the responsible NLP checklist. All AI-assisted outputs used in the paper, code, and bibliography were reviewed, edited, and verified by the authors, who remain responsible for the correctness of the submission.

\subsection{Case Studies}

We supplement the quantitative analyses of Section~\ref{sec:analysis} with three case studies drawn from the diagnostic subset, summarised in the box below. In all three cases, a positive $\Delta_j$ indicates that the item pushes the posterior toward the option LEAP currently places on top, and a negative $\Delta_j$ indicates that it pulls the posterior away. We then expand one continuous case to show the audit trace exposed by LEAP: prior construction, local likelihood elicitation, safeguard decisions, posterior parameters, and leave-one-out contributions.

\begin{tcolorbox}[casebox, title={Case studies from the diagnostic subset}]
\small
\textbf{Case 1. Confident Monolithic error, calibrated LEAP.}\\
\textbf{Question.} Identify a cathedral from clues: 13th-century construction, 17th-century damage, effigies and grave slabs, riverbank location.\\
\textbf{Ground truth.} \textbf{A.} St.~Canice's Cathedral.\\[2pt]
\begin{tabular}{@{}l@{\hspace{0.5em}}l@{\hspace{0.5em}}l@{}}
\emph{Monolithic} & B. Christ Church $\;(p\!=\!0.85)$ & incorrect \\
LEAP   & A. St.~Canice's $\;(p\!=\!0.54)$ & correct \\
\end{tabular}\\[2pt]
\textbf{Per-evidence contributions (top 3 of 5):}\\[1pt]
\begin{tabular}{@{}p{0.62\linewidth} r@{}}
\quad Grave slabs identification & $\Delta\!=\!+0.249$ \\
\quad Christ Church search       & $\Delta\!=\!-0.248$ \\
\quad Maps summary               & $\Delta\!=\!+0.190$ \\
\end{tabular}\\[2pt]
\textbf{Takeaway.} \emph{Monolithic} concentrates probability on the wrong option; LEAP places the correct option on top with conservative confidence and exposes which evidence items drove the decision.

\par\medskip\hrule\medskip

\textbf{Case 2. Confidently correct LEAP under noisy evidence.}\\
\textbf{Question.} Winner of the 2026 Rugby Europe Championship (Georgia, Portugal, Romania, Spain, or Other).\\
\textbf{Ground truth.} \textbf{A.} Georgia.\\[2pt]
\begin{tabular}{@{}l@{\hspace{0.5em}}l@{\hspace{0.5em}}l@{}}
\emph{Monolithic} & B. Portugal $\;(p\!=\!0.33)$ & incorrect \\
LEAP   & A. Georgia $\;(p\!=\!0.80)$ & correct \\
\end{tabular}\\[2pt]
\textbf{Per-evidence contributions (top 3 of 5):}\\[1pt]
\begin{tabular}{@{}p{0.62\linewidth} r@{}}
\quad Official tournament page  & $\Delta\!=\!+0.361$ \\
\quad Semifinalist bracket      & $\Delta\!=\!-0.081$ \\
\quad Georgia semifinal result  & $\Delta\!=\!+0.027$ \\
\end{tabular}\\[2pt]
\textbf{Takeaway.} \emph{Monolithic} distributes mass over four nearly equal options. LEAP concentrates on the correct option through one strong supporting item, even with a second item that pulls in the opposite direction. The two minor items contribute negligibly to the posterior.

\par\medskip\hrule\medskip

\textbf{Case 3. Honest uncertainty under absent evidence.}\\
\textbf{Question.} Identify the publication date of a piece of writing (four candidate dates close in time).\\
\textbf{Ground truth.} \textbf{A.} January 2, 1975.\\[2pt]
\emph{Monolithic} predicts B. Jan 2, 1974 with $p\!=\!0.63$ (incorrect); LEAP returns the uniform distribution $p\!=\!0.25$ over the four options (no commitment).\\[2pt]
\textbf{Per-evidence contributions:} none retained ($|\mathcal{R}|\!=\!0$ after dependency clustering).\\[2pt]
\textbf{Takeaway.} When no usable evidence is collected, LEAP returns the uniform prior rather than committing to a guess, whereas \emph{Monolithic} produces a confident but incorrect prediction. The uniform forecast is poorly resolved but well calibrated; the Monolithic forecast is sharper but concentrates probability on the wrong option. This is exactly the calibration tradeoff discussed in Section~\ref{sec:analysis}.
\end{tcolorbox}

\paragraph{Expanded audit trace.}
The following continuous forecasting case illustrates what a reviewer can inspect beyond the final prediction. The task asks for the daily high of AAPL on January 23, 2026, with ground truth 249.41 USD. \emph{Monolithic} predicts 249.0 with an 80\% interval of [245.5, 253.5]. LEAP first constructs a data-derived prior from the latest available seven daily highs,
\[
\begin{aligned}
\relax[&261.55, 261.56, 260.78, 258.64,\\
 &254.53, 251.30, 250.74],
\end{aligned}
\]
giving $P_0=\mathcal{N}(250.74, 7.5222^2)$. This prior is not an LLM forecast; it is the base distribution before evidence likelihoods are added.

\begin{table}[h]
\centering
\small
\setlength{\tabcolsep}{3pt}
\begin{tabularx}{\linewidth}{@{}l r r X@{}}
\toprule
\textbf{Item} & $\boldsymbol{\mu_i}$ & $\boldsymbol{\sigma_i}$ & \textbf{Local LLM rationale} \\
\midrule
E1 & 250.74 & 12.54 & Previous daily high directly matches the target scale; uncertainty is set at roughly 5\% of the value. \\
E5 & 247.65 & 12.40 & Nasdaq quote gives a recent AAPL opening value, informative but not identical to the next daily high. \\
E12 & 190.00 & 47.50 & Trefis reports a 190 USD analyst target; the model reads it as target-scale evidence. \\
Other items & 250.74 & 300--1000 & Headlines, broad indices, or macro context provide only weak proxies, so the local likelihood is intentionally diffuse. \\
\bottomrule
\end{tabularx}
\caption{Representative local likelihood outputs in the expanded continuous case. Each row is produced from one isolated evidence item; the LLM does not see the other evidence items or the posterior.}
\label{tab:case-expanded-likelihoods}
\end{table}

The safeguard step rejects E12 as an elicitation error because
\[
z = |190.00 - 250.74| / 7.5222 = 8.07 > 4.
\]
This catches a semantic mismatch: an analyst target price is not a direct estimate of the next daily high. The remaining 17 representatives enter the Gaussian update. Their total precision is
\[
\tau_{\mathrm{post}} = 0.017672 + 0.013069 = 0.030741,
\]
and the posterior is
\[
P(\theta \mid \mathcal{R})=\mathcal{N}(250.157, 5.7031^2).
\]
Most posterior precision comes from the prior, E1, and E5, with approximate weights 57.5\%, 20.7\%, and 21.2\%, respectively; all other retained items together contribute less than 1\%.

The leave-one-out trace identifies how the final mean changes when each retained item is removed: E5 contributes $\Delta=-0.6727$, E1 contributes $\Delta=+0.1520$, E8 contributes $\Delta=+0.0680$, and the remaining items are near zero. Thus the final forecast is accompanied by the local LLM interpretations, the deterministic posterior calculation, the rejected outlier, and reproducible evidence-level contributions.

\end{document}